\pdfoutput=1 
\documentclass[pmlr]{jmlr}

\usepackage{longtable}
\usepackage{placeins}
\usepackage{caption}

\usepackage{booktabs}
\usepackage[load-configurations=version-1]{siunitx} 
\usepackage{comment}

\makeatletter
\def\set@curr@file#1{\def\@curr@file{#1}} 
\makeatother

\theorembodyfont{\upshape}
\theoremheaderfont{\scshape}
\theorempostheader{:}
\theoremsep{\newline}

\jmlrvolume{126}
\jmlryear{2020}
\jmlrworkshop{Machine Learning for Healthcare}

\graphicspath{ {./figures/} }
\title[Using deep networks for scientific discovery in physiological signals]{Using deep networks for scientific discovery in \\ physiological signals}

\author{\Name{Tom Beer*}
       \Email{tom1beer@gmail.com}\\ 
       \addr Technion - Israel Institute of Technology\\
       \AND
       \Name{Bar Eini-Porat*}
       \Email{briany202@gmail.com}\\ 
       \addr Technion - Israel Institute of Technology\\
       \AND
       \Name{Sebastian Goodfellow}
       \Email{sebi.goodfellow@utoronto.ca}\\ 
       \addr University of Toronto\\
       \AND
       \Name{Danny Eytan}
       \Email{danny.eytan@technion.ac.il}\\ 
       \addr Technion - Israel Institute of Technology and University of Toronto\\
       \AND
       \Name{Uri Shalit}
       \Email{urishalit@technion.ac.il}\\ 
       \addr Technion - Israel Institute of Technology\\
       \AND
       * Equal contribution, alphabetical order 
       }

\begin{document}

\maketitle

\begin{abstract}
Deep neural networks (DNN) have shown remarkable success in the classification of physiological signals. In this study we propose a method for examining to what extent does a DNN's performance rely on rediscovering existing features of the signals, as opposed to discovering genuinely new features. Moreover, we offer a novel method of ``removing'' a hand-engineered feature from the network's hypothesis space, thus forcing it to try and learn representations which are different from known ones, as a method of scientific exploration. We then build on existing work in the field of interpretability, specifically  class activation maps, to try and infer what new features the network has learned. We demonstrate this approach using ECG and EEG signals. With respect to ECG signals we show that for the specific task of classifying atrial fibrillation, DNNs are likely rediscovering known features. We also show how our method could be used to discover new features, by selectively removing some ECG features and ``rediscovering'' them. We further examine how could our method be used as a tool for examining scientific hypotheses. We simulate this scenario by looking into the importance of eye movements in classifying sleep from EEG. We show that our tool can successfully focus a researcher's attention by bringing to light patterns in the data that would be hidden otherwise. 
\end{abstract}

\section{Introduction}

In the last decade, we have seen tremendous breakthroughs in using deep-learning based data representations \citep{lecun2015deep}. These representations have proven to be highly effective in complex tasks such as computer vision, natural language processing, reinforcement learning, and also in healthcare \citep{esteva2019guide}. The key advantage of deep neural networks compared to previous learning approaches is that their inductive bias and expressive structure makes them superb feature extractors \citep{sharif2014cnn}, often replacing the need for hand-engineering features.

In many cases these hand-engineered features represent the accumulation of decades of scientific research and understanding of the underlying physical processes. For example, when using electrocardiogram (ECG) signals for identifying a specific arrhythmia named atrial fibrillation (AF), there are sets of features derived from understanding of the underlying pathological process, based on years of observations. More specifically these features are focused on quantifying the irregularity of the heart beats and identification of what are known as ``P waves'': distinct voltage fluctuations that represent atrial depolarization which are usually absent in atrial fibrillation. When applying deep-learning based approaches to the same task of identifying AF, we see that these approaches perform similarly in terms of accuracy without us encoding the scientific knowledge acquired about such signals. How do deep networks do it? In this paper we provide a method whose first goal is to try and uncover to what degree do the networks re-discover known features, and to what degree do they discover genuinely new aspects of the signal. If they exist, these newly discovered features could hopefully be used to further our scientific understanding of the mechanisms underlying signal generation. \emph{Moreover, we hypothesized that it is possible to ``push'' deep networks towards discovering new features}, using a method we describe below. 

These tasks are far from easy, since deep neural networks are famously ``black-boxes'' that do not explain their predictions in a way that humans can easily understand and derive insight from. We propose a new method which is focused on explaining what features a deep neural network uses, \emph{shown in relation to known features}. 
To do this, we offer a novel method of ``removing'' a hand-engineered feature from the network's hypothesis space, thus forcing it to try and learn representations which are different from known ones. We then build on existing work in the field of interpretability, specifically  class activation maps \citep{zhou2016learning}, to try and infer what new features the network has learned. These new features in turn might offer insights into new scientific knowledge regarding the signal. 

For example, we consider the task of identifying atrial fibrillation from an input of ECG signals: We extract hand-engineered features from the signal. Then, we train a neural network to predict the AF label, while adding a penalty to one of the network's layers, forcing it to be as statistically independent as possible from the hand-engineered features, while performing the task. This is achieved by using the Hilbert-Schmidt Independence Criterion (HSIC), see Section \ref{Methods}. We then examine how well does the network perform under this independence constraint, and examine the activation maps to see what new features does the network learn when it is forced to abandon previously known features.

In this case we find that when one type of features is removed, namely RR-peaks, then the network focuses on P-waves. We further find that once we force the network to use neither RR-peaks nor P waves, its performance plummets to almost chance. This implies that in the case of ECG signals and atrial fibrillation, long-known hand-engineered features indeed capture almost all of the signal; deep neural nets seem to rediscover these features, but do not discover new ones, as far as we can tell. 

In addition, we examine a different type of physiological signal - the Electroencephalograms (EEG) signals of subjects in various stages of sleep and show that our method can be used to validate hypotheses regarding putative features and their underlying generative processes that can serve to differentiate between sleep stages.  



\subsection*{Generalizable Insights about Machine Learning in the Context of Healthcare}
We show how deep networks' success in tasks involving physiological signals can be leveraged towards the difficult task of scientific discovery. Along the way, we find evidence showing that state-of-the-art networks for labeling ECG signals as having AF seem to be rediscovering well-known features, while not adding much beyond these features, while on EEG signals we show that hypotheses regarding the presence of  eye movements signatures in the EEG signal can be corroborated using our  method and these can serve to help classify sleep stages. Finally, we also provide a technical tool to ``remove'' a predefined signal from the hypothesis space of a deep neural network. This tool might be useful for tasks beyond those we consider here: ensuring robustness to certain distribution changes, or understanding the working of the network from a specific angle.

\section{Method} \label{Methods}
We propose a framework for learning a deep representation of a physiological signal, such that the learned representation does not include components of the signal associated with predefined, hand-engineered features. Our code is available at \href{https://github.com/shalit-lab/deep-scientific-discovery}{github.com/shalit-lab/deep-scientific-discovery}.

We assume we have access to a set of $n$ samples of the form $(x_1,f_1,y_1), \ldots , (x_n,f_n,y_n)$, where: $x_i$ are raw signals, for example ECG signals; the vectors $f_i$ are hand-engineered features, calculated as a deterministic function of $x_i$, and typically of much lower dimension than $x_i$; and, $y_i \in \{1,\ldots,k\}$ are discrete labels, for example whether the signal is of atrial fibrillation in ECG, or REM sleep stage vs. Non-REM sleep in EEG signals. In the example of ECG signals, $f_i$ might be P-wave features such as the maximum, standard deviation and energy of the amplitude \citep{goodfellow2018atrial}, see \ref{ecg-features}. In the example of EEG signals, $f_i$ might be relative bandpowers for frequency bands such as Delta, Alpha and Beta \citep{al2014methods}, see \ref{eeg-features}.

The main goal of the architecture and loss function we propose below are to learn DNN features (denoted $g_i$) which are on the one hand informative about the labels $y_i$, but on the other hand contain as little information as possible about the hand engineered features $f_i$. We do this by combining two elements: a term added to the objective function which encourages statistical independence between the $f_i$ and $g_i$, and an architecture which encourages $g_i$ to be non-redundant with respect to $f_i$. 

The measure we choose to enforce independence between the hand-engineered features $f_i$ and the DNN's representation $g_i$ is the Hilbert-Schmidt Independence Criterion (HSIC), which is based on the eigenspectrum of covariance operators  in  reproducing  kernel  Hilbert  spaces  \citep{gretton2005measuring}.

There are various other measures of statistical independence; the advantage of HSIC is that it is non-parametric, unlike mutual information, and thus fit for samples not following a prescribed distributional form. In addition, it does not require training an additional model (i.e. an inference network for variational approximation or an adversarial network).
 
HSIC can be thought of as a non-linear extension of the cross-covariance between two random variables. Unlike the cross-covariance, the HSIC between two random variables $X$ and $Y$ equals $0$ if \emph{and only if} $X$ is independent of $Y$ (under certain regularity conditions).
There are of course other ways of measuring and encouraging statistical independence, for example using mutual information or adversarial networks. However, mutual information requires  parametric assumptions about the distribution, which would be hard to make here, or alternately training a variational network as in \citet{kim2019learning}. The advantage of HSIC is that it is non-parametric, and thus fit for samples not following a prescribed distributional form. In addition, it does not require training an additional model such as an inference network for variational approximation of MI, or an adversarial network as in \citet{ganin2016domain} and followup work.

The HSIC term is calculated in the following manner: For reproducing kernel Hilbert spaces $\mathcal{F, G}$ with universal kernels $k$, $l$ we compute the kernel matrices $K_{ij} = k(f_i,f_j)$ and $L_{ij} = l(g_i,g_j)$. In our case both $k$ and $l$ were selected to be Gaussian kernels:
$K_{i,j} = e^{\frac{{\lVert f_i-f_j \rVert}^2_2}{\sigma^2}}$,
with $L_{i,j}$ computed in a similar manner. Then the scaled Hilbert-Schmidt norm of their cross covariance matrix is calculated:

\begin{equation*}
    \widehat{HSIC}(\{(f_i,g_j\}^n_{i=1}; \mathcal{F, G} ) = \frac{1}{(n-1)^2}  \cdot \textbf{tr}(KHLH),
\end{equation*}
where $H_{ij}=\delta_{i,j}-\frac{1}{n}$ is a centering matrix. For each kernel matrix, the bandwidth $\sigma$ is set to be the median pairwise distance of the data points, a heuristic described in \citet{mooij2009regression}. Since $\{f_i\}^n_{i=1}$ is a fixed representation space, its median pairwise distance can be computed once before training. However $\{g_i\}^n_{i=1}$ changes every training step, so its bandwidth is updated accordingly at every training step based on a moving average between the current median pairwise distance and the previously set bandwidth.

The hand-engineered features $f$ are usually highly predictive for the selected task. In order to further encourage the DNN latent representation $g$ to be distinct from $f$, we concatenate the two representations before passing them on to the final layer of the network. We do this to prevent, as much as possible, from the network representation $g$ to try and replicate $f$, even when under the HSIC constraint. We have found that it is not possible to achieve both reasonable classification accuracy on the original task and sufficient independence from the external representation without this concatenation step.

The overall flow is as follows: The raw signal inputs $x_i$ are passed through a convolutional neural network.  Specifically in our experiments we use a 12-layer convolutional neural network inspired by WaveNet \citep{oord2016wavenet}. The same signals are also passed to a hand-crafted feature extractor which outputs the features $f_i$. Each channel of the CNN's output is reduced to a single number by global average pooling, resulting in a feature vector of size 512, which we considered to be the DNN's latent representation of the raw signal and denote as $g_i$. This latent  representation is concatenated along with the external features, yielding a unified representation $[f_i, g_i]$. The classifier's output $\hat{y}_i$ are then computed by a softmax of a fully connected linear layer from this concatenated representation.
The loss function of the network's output is given by:
\begin{equation}
    \lambda \text{HSIC}\left([f_1,\ldots f_n];[g_1,\ldots,g_n]\right) + \sum_{i=1}^n \text{CrossEntropy}\left(\hat{y}_i,y_i\right),
\end{equation}
where $\lambda$ is a hyperparameter controlling the degree of independence we induce between the DNN representation $g_i$ and the hand-engineered representations $f_i$. This parameter is tuned to ensure both high performance on the original task and sufficient independence relative to the external features. See section \ref{Evaluation Approach} and figure \ref{fig:rr_model_selection} for more details.


\begin{figure}[htbp]
  \centering 
  \includegraphics[width=0.7\textwidth]{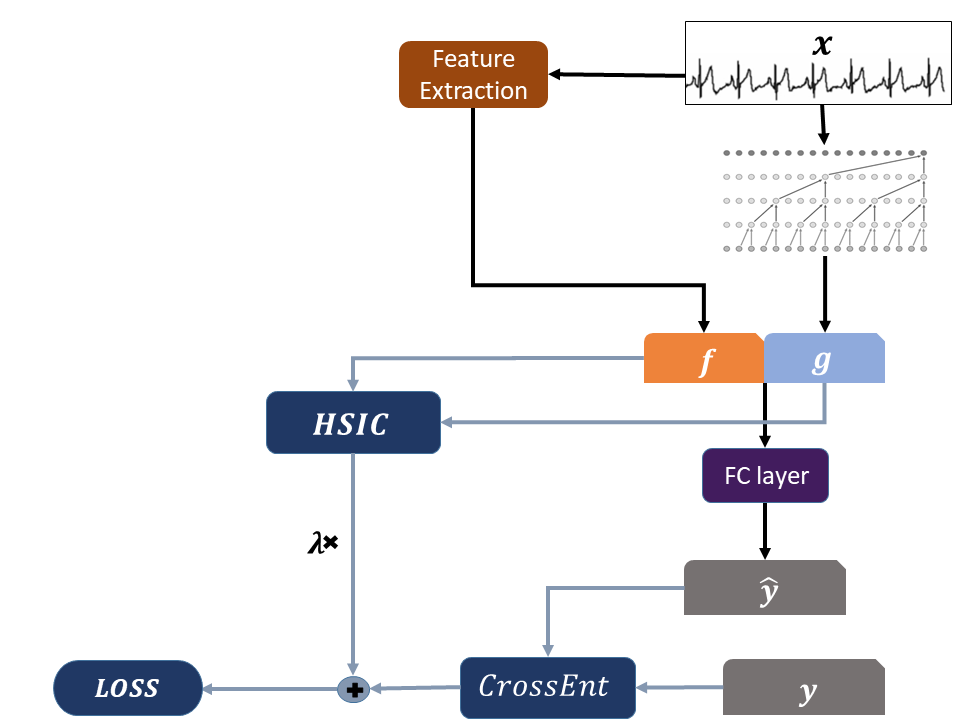} 
  \caption{Network flow and objective computation: Input signals $x$ are passed through the DNN, yielding  $g$, the latent representation. $g$ is concatenated with extracted feature set $f$ and passed through the FC layer, generating the final class logits $\hat{y}$. Then, HSIC is calculated for $g$ and feature set $f$.
  Finally, the weighted HSIC loss is added to the cross entropy loss of $\hat{y}$ and true labels $y$. Note that the black arrows refer to the internal network flow and the blue arrows refer to the calculation of the loss.}
  \label{fig:HSICFlow} 
\end{figure} 

\vspace{1em}

We chose the WaveNet architecture as similar models were used both for arrhythmia detection from ECG signals \citep{goodfellow2018towards} and for sleep stages classification from EEG signals \citep{fernandez2018convolutional}, which are the two tasks we focus on in our experiments.

Figure \ref{fig:HSICFlow} schematically summarizes the described method.
 


\subsection{Visualization of activation}\label{methods:vis}
Once we learn a representation $g$, we wish to understand its relation to the signal $x$, trying to answer the question ``which aspects of the signal does $g$ capture and focus on?''.
We address this challenge using the idea of class activation maps, CAMs \citep{zhou2016learning}. CAMs are a technique for  visualising spatially localized class-relevant information. They are obtained by a matrix multiplication between a CNN's activations (before global average pooling) and the weights of the fully connected layer. Here we follow the usage of \citet{goodfellow2018towards} who applied CAMs to  physiological time series data, thus obtaining time-localized information about class activations. 

For a physiological signal the output of CAM is, for each sample, a time series showing what parts of the signal produce strong activations per a given class. We (1) align these activation profiles around a meaningful physiological landmark, e.g. an R-peak in ECG signals or a rapid eye movement in EEG signals, (2) normalize the activation profiles to $[0,1]$, and (3) average them over the entire test set. See section \ref{ecg-features} and \ref{eeg-features} for details about RR-peak and rapid eye movement alignment. 
We note that when the mere existence of an event is significant, we would expect CAM peak over the event start. This high activation over the event's start is interpreted as importance for the task.




\section{Experimental Setup}
We evaluate our proposed method with two experimental tasks using different physiological signals: arrhythmia detection in electrocardiogram signals (ECG), and REM Non-REM classification on brain-derived signals, the encephalogram (EEG). 

In the ECG case we experiment with combinations of two feature sets (RR and P-Wave), and use a single type of temporal landmark to align the Class Activation Maps: the R-peak. In the EEG case we use one feature set (frequency features), and experiment with two different temporal landmarks for aligning the Class Activation Maps: slow waves and rapid eye movements.

\subsection{ECG Experiment}
\subsubsection{Motivation}
The first experiment chosen to demonstrate our proposed approach is the classification of disturbances in the electrical impulse generation of the heart leading to abnormal heart rates - arrhythmias - using  single lead ECG waveforms. The ECG is a tracing of the electrical activity that is taking place within the heart as recorded by electrodes (leads) placed on the skin. Under normal circumstances, an electrical impulse will travel from the sinoatrial node (SA-node), spread across the atrium to generate a distinct deflection named P-wave, to the atrioventricular node and through the ventricular septum of the heart. This electrical impulse causes the four chambers of the heart to contract and relax in a coordinated fashion with the spread of the signal in the ventricles identified by a distinct series of deflections called the QRS complex. Special attention is devoted to the prominent peak of the QRS complex, known as the R-peak. Studying these electrical signals allows detection of abnormalities in the rhythm, rate or even blood supply to the heart. In this study we chose to focus on the classification of a common arrhythmia named atrial fibrillation (AF), the most common serious abnormal heart rhythm \citep{censi2016p}. In AF, the normal regular electrical impulses generated by the SA-node in the right atrium of the heart (those that generate the P-waves on the ECG) are overwhelmed by disorganized electrical impulses leading to irregular conduction of ventricular impulses that generate the heartbeat. AF is characterized by rapid and chaotic beating of the atrial chambers of the heart and loss of the P-wave along with irregular R-peaks. Thus, most classic ML models for detecting AF are based on features that try to capture the loss of the regularity of the heart rhythm and the absence of P-waves. 

As we explain below, we will experiment with two different types of hand-engineered features $f$: one relates to the P-wave, and the other relates to the R-peaks. We provide the model with one feature set (thus effectively pushing it to extract features that are independent of that feature type) and ask if we can then discover the other feature set by examining the class activation maps of the model. We denote RR-model as the model provided with the R-peak feature set, P-wave model as the model that is provided the the P-wave feature set and All model is the model provided with both feature sets. 

The main reason for selecting this domain for the design, development and demonstration of our framework is its long history in the healthcare, signal processing and machine learning communities. It has been studied for over a century by experts and doctors. In the last decades, ECG signals were analyzed by signal processing engineers; and in the last couple of years, machine learning specialists have come up with data driven approaches to arrhythmia detection. All in all, the scientific community holds extensive medical and engineering knowledge about this task. This extensive knowledge will serve as ground truth that is critical for evaluation of our approach.

\subsubsection{Features and alignment} \label{ecg-features}
The features were extracted using the method of \citet{goodfellow2018atrial}. Of note, other top scoring groups at the 2017 Physionet challenge \citep{clifford2017af} relied on hand-engineered featured-based models demonstrating F1 scores of 0.83, similar to the SOTA in this field as tested on this specific dataset. 

\begin{itemize}
    \item {\textbf{RR Features:}}
    The RR is the sequence of time intervals between successive R-peaks. For this analysis, the following features reflecting heart rate variability were used: Median, standard deviation, root mean square, multiscale entropy \citep{costa2005multiscale}, minimum and maximum of the RR sequence, and PNN20, PNN50 which denote the fraction of RR intervals where the change in consecutive RR intervals exceeds 20 and 50 milliseconds, respectively. The total dimension of this feature set is 8.
    \item {\textbf{P-Wave Features:}}
    The P-wave window is defined as the interval ranging from 250 ms to 100 ms prior to each R-peak, where the voltage deflection named the P-wave can be found when it exists. From the set of P-wave windows, the following features were extracted: Maximum, standard deviation and energy of the amplitude, median and standard deviation of the correlation coefficient matrix, median Higuchi fractal dimension and the time of the maximal amplitude. The total dimension of this feature set is 7.
\end{itemize}

Finally, as explained in section \ref{methods:vis}, in our experiments we visualize the class activation maps by aligning them to physiological landmarks in the signal. For the ECG signals, we use the R-peak as this landmark.

\subsubsection{Data and task definition}

The dataset used for the ECG experiment was originally published for the 2017 Physionet challenge \citep{clifford2017af}.
The database consists of single-channel ECG waveforms that were acquired by patients using AliveCor\,\textsuperscript{\tiny\textregistered}’s ECG device, and ranged in duration from 9 seconds to 60 seconds.

The main task we consider for this experiments is a binary classification task: Detection of atrial fibrillation. The negative class (NAF) consists of normal sinus rhythm which is the default heart rhythm, along with many different heart arrhythmias such as ventricular tachycardia, atrial flutter, ventricular bigeminy, and ventricular trigeminy. 

\begin{table}[h!]
\centering
\begin{tabular}{c c c c c c} 
 \hline
&&\multicolumn{3}{c}{Duration (seconds)} \\
\cline{3-6}
 Label & Count & Mean & Std & Max & Min \\ [0.5ex] 
 \hline
 Atrial fibrillation & 771 & 31.6 & 12.5 & 60 & 10 \\
 NAF & 7,711 & 32.6 & 10.7 & 61 & 9 \\
 Total & 8482 & 32.5 & 10.9 & 61 & 9 \\ [1ex] 
 \hline
\end{tabular}
\caption{ECG Dataset Summary}
\label{table: ECG Dataset}
\end{table}

Examples with the label ``noisy'' (given in the original dataset) were filtered out, leaving 8482 viable records. The signals were downsampled from 300Hz to 90Hz, and were zero padded to a length of 5400 samples (60 seconds). The dataset is summarized in Table \ref{table: ECG Dataset}.



\subsubsection{Training Framework}
To obtain a balanced classification task, the positive class (atrial fibrillation) was upsampled to 50:50 ratio. The dataset was further split into a training, validation and test sets (70\%/15\%/15\% split), preserving class proportions. The model was trained for 100 epochs with a batch size of 32, and a cosine annealed learning rate ranging between $10^{-3}$ and $10^{-5}$. Optimization was preformed with ADAM optimizer \citep{kingma2014adam} and cross-entropy loss, with the additional HSIC term calculated as in \citet{greenfeld2019robust}, and $\lambda$ set to 500. For more details see appendix \ref{appendix:main_details}.


\subsection{EEG Experiment}
\subsubsection{Motivation}
The second experiment focuses on sleep stages in EEG signals. 
Sleep is a rapidly reversible state of reduced responsiveness, motor activity, and metabolism \citep{siegel2009sleep}. It is classically divided into four distinct \emph{sleep stages}: Stage N1 (formerly stage 1 sleep), stage N2 (formerly stage 2 sleep), stage N3 (formerly stages 3 and 4 sleep), and stage R sleep (formerly stage REM sleep) as defined by the American Academy of Sleep Medicine (AASM) \citep{berry2012aasm}. Sleep staging relies on several signals: those generated from the cerebral cortex (EEG), eye movements (electro-oculogram, EOG), chin electromyographic activity (EMG), ECG and additional signals collected to identify sleep apnea and disturbances. The EEG is the main source for sleep stage classification, with each stage defined by the frequency content and distinct features with characteristic signatures such as sleep spindles and K-complexes. Machine learning and specifically deep learning has shown promise in classifying sleep stages from EEG signals \citep{aboalayon2016sleep, roy2019deep}. 
In this work, we focus on REM (stage R) vs. Non-REM (stages N1-3) classification. 


\subsubsection{Data and task definition}
This experiment represents a rather different use case than the previous one: here we simulate the case where the researcher has a hypothesis about a novel component of the signal that she believes might be useful. In this scenario, the researcher can use our proposed framework to ``de-noise'' the DNN's class activation maps by removing known aspects of the signal, hopefully allowing better focus on the hypothesized novel component.


Specifically we examine the following scenario: Suppose that the researcher suspects that signatures of rapid-eye-movement events are present in the EEG trace and are significant for REM/NREM classification. She may train a model to ``ignore'' other known EEG features that support REM/NREM classification and examine the residual activation over EEG traces surrounding rapid-eye-movement events (identified using the other traces in the sleep study, namely the EOG), in order to determine whether they are relevant for REM detection task from EEG.






The dataset is taken from the Sleep Heart Health Study (SHHS) \citep{quan1997sleep}. We used data from first round of recording (SHHS1), which includes two bipolar EEG channels (C4/A1 and C3/A2) recorded at 125Hz, which were recorded along with other physiological signals such as EOG and EMG. Each record was manually scored by a single technician on 30-second epochs according to \citet{rechtschaffen1971manual}. Due to memory and computational constrains we subsample 5\% of each patient's sleep records, yielding a dataset of 187358 labelled examples. The EEG signals were downsampled to 80Hz which we verified did not degrade the classification performance of any of the models we used.

Table \ref{table: EEG Dataset} summarizes the key quantities in the SHHS dataset.

\begin{table}[h!]
\centering
\begin{tabular}{c c c c c c} 
 \hline
\cline{3-6}
 Label & Number of sleep epochs \\ [0.5ex] 
 \hline
 Non-REM & 149,286  \\
 REM & 38,072 \\
 Total & 187,358 \\ [1ex] 
 \hline
\end{tabular}
\caption{EEG Dataset Summary}
\label{table: EEG Dataset}
\end{table}


\subsubsection{Features and alignment}\label{eeg-features}
\begin{itemize}
    \item {\textbf{Frequency features:}}
    Relative bandpowers were extracted from Welch's periodogram for the following frequency bands: Delta (0.5-4Hz), Theta (4-8Hz), Alpha (8-12Hz) and Beta (12-30Hz). The total dimension of the frequency feature set is 4. 
\end{itemize}

The following were identified as potential temporal landmarks of interest:
\begin{itemize}

    \item {\textbf{Slow waves:}}
    We utilize the method proposed by \citet{carrier2011sleep}, \citet{massimini2004sleep} as implemented in \citet{raphael_vallat_2020_3646596} to extract slow wave locations. 
    \item {\textbf{Rapid Eye Movements}} Bursts of rapid eye movements (REMs) are a defining feature of REM sleep. They are usually detected using the EOG channel. Rapid eye movements' locations were extracted directly from the corresponding EOG channel as implemented in \citet{raphael_vallat_2020_3646596}.

\end{itemize}

In this experiment, we only used the frequency features: We denote Frequency-model as the model provided with the frequency feature set.
As detailed above, to align the class activation maps in the EEG experiment we use the timing of the rapid eye movement as determined from the EOG signal. Note that while in the ECG experiment the aligning landmark is part of the signal used by our method, in the EEG experiment one of the two aligning landmarks comes from a different signal, the EOG, which is not directly available to our model.

\subsubsection{Training Framework}
For the purpose of the REM Non-REM task, the wake records were dropped from the dataset and all the N-REM classes (N1-3) were unified to one large NREM class. The records were further split into training, validation and test sets (50\%/20\%/30\% split) while preserving class proportions.

Optimization was preformed by ADAM optimizer \citep{kingma2014adam} on the objective function defined in section \ref{Methods}, with the additional HSIC term calculated as in \citet{greenfeld2019robust}. The HSIC weighting, $\lambda$, is set at each batch to keep a constant proportion of $0.25$ between the classification loss and the independence loss. For more details see appendix \ref{appendix:main_details}.

\section{Results}
Our major results revolve around examining the class activation maps (CAM) of our model in two tasks: one using ECG signals, and one using EEG signals.

\subsection{Evaluation} \label{Evaluation Approach}

In each of the experiments, before we address our main task, we run a series of tests to verify that (1) All models and representations (both DNNs and hand-engineered features) should contain meaningful information regarding the classification task. (2) Training with the CrossEntropy + HSIC loss has succeeded in achieving independence between the hand-engineered features and internal network representation, while (3) retaining some information about the label within the DNN representation. 
We thus have three auxiliary tasks, whose aim is to boost our confidence in the claims made in the main task. 
The models for these tasks are detailed in appendix \ref{appedix:aux_details}.

\begin{itemize}
    \item{\textbf{Relevance of hand-engineered features $f$ (\emph{Relevance}):}}
    Any set of hand engineered features used for this framework should contain information with respect to the main classification task. To measure this we train a vanilla neural network predicting the label $y$ directly from the hand-engineered features $f$ and measure held-out accuracy.
   
    \item{\textbf{Testing for independence between hand-engineered features and DNN features (\emph{Independence}):}}
    The HSIC term in our loss function is meant to induce statistical independence between the hand-engineered features $f$ and the DNN features $g$. We validate that this is indeed achieved by checking whether one can predict the features $f$ when given the features $g$ as input. We do this by training a multi-task DNN whose input is $g$ and output labels are the entries of $f$. We then measure the held-out squared correlation $R^2$, averaged across the dimensions of $f$.
    An average $R^2$ value close to 0 indicates a high degree of independence. As a further reference point, this value is compared to the $R^2$ value obtained using an network with the same architecture as ours but without the HSIC term (equivalent to setting $\lambda=0$), which we call the Baseline Model.
    
    \item{\textbf{Label information in latent representation (\emph{Rep2Label}):}}
    To confirm that the obtained representation $g$ holds useful information for the task in question, we run the following evaluation:
    Our method is trained on a sample set $train_1$. Then we apply the representation function $g$ to a separate sample set $train_2$. We fit a vanilla 3-layer network, which we call Rep2Label, predicting $y$ from $g$ on the training set $train_2$. Finally, we evaluate the accuracy of Rep2Label on a held-out test set.
    An accuracy greater then chance indicates that the latent representation still holds valuable information about the label and can be analyzed for our goal of scientific discovery.
    
\end{itemize}

  
Finally, for our main task we train our model with different choices of the hand-engineered features $f$, using the validation set for hyperparamter tuning, and examining the temporal Class Activation Maps on the held-out test set as described in Section \ref{Methods}.
We note that since the existence or absence of an event is meaningful, we would expect a peak in class activation over the event start. For example, in the ECG experiment CAM peaks are expected over P-wave or QRS start, whereas in EEG experiment they would be expected over slow-wave, spindle or rapid-eye-movement starts.



\subsection{Results: ECG Experiment} 
As detailed in \ref{ecg-features}, in this experiment we use two different sets of hand-engineered features: those based on R-peaks and those based on P-Waves. We therefore have 4 models: a Baseline Model which does not use these features at all, an RR Model and a P-wave Model, each with the corresponding features, and a model which uses both sets of features which we call All Model. 
We first examine the relevance of the two sets of hand-engineered features used in this task, the RR and the P-wave sets. In Table \ref{table:ECG relevance} we see that both lead to reasonably high accuracy when used as input to a simple neural network for predicting AF, with RR being more informative than the P-waves. We note that since the labels are balanced, chance performance here is $50\%$ accuracy. 

We proceed to examine whether HSIC indeed manages to induce independence between the learned representations $g$ and the hand engineered features $f$. In Table \ref{table:ECG results} we note that for the baseline model, the average $R^2$ when trying to predict $f$ from $g$ is 0.51 for predicting RR features and 0.1 for predicting P-Wave features. Contrary to that, we see that in the models with the HSIC penalty the corresponding hand-engineered features are essentially unpredictable from the learned representation $g$, sometimes doing worse than just predicting the mean, leading to negative $R^2$ values. We take this to be an indication that our method is successful in learning representations that are indeed independent of the  hand-engineered features used in each model.

We then proceed to examine the third auxiliary task, Rep2Label, whose goal is to examine whether the learned representations still carry information about the target label. We see that while the baseline model is highly predictive of the label (on par with the best hand-engineered features), once we remove either the P-wave or the RR features, the learned representation carries much less information about the labels, with performance dropping precipitously. When we remove \emph{both} feature sets, performance of using solely the representation is not statistically significantly different from chance performance ($n=620, p=0.2$). The significant drop in classification accuracy of the models constrained by only one of the feature sets can be explained by the fact that the two feature sets are highly correlated; constraining one set causes the model to ignore some information reflected by the other set.


Finally, we examine the accuracy and F1 scores of the full models, i.e. using the constrained representation $g$ concatenated with the hand-engineered features $f$. These models achieve slightly better performance than using merely the hand-engineered features.

We therefore come to the conclusion that state-of-the-art type DNNs for predicting AF from ECG might not seem to learn much beyond the information contained in the combined set of RR and P-Wave hand-engineered features. Remarkably, they seem to have ``rediscovered'' without explicit guidance the best features as discovered over years of work by human experts.



\begin{table}[h!]
\centering
\begin{tabular}{c c c c c c} 
 \hline
 Feature Set & Accuracy & F1 \\[0.5ex] 
 \hline
RR feature set & 93.9\% & 0.91 \\ 
P-Wave feature set & 87.3\% & 0.86\\
All feature set & 95.5\% & 0.95\\

 \hline
\end{tabular}
\caption{Relevance task performance for the two hand engineered feature sets.}
\label{table:ECG relevance}
\end{table}

\begin{table}[h!]
\centering
\begin{tabular}{c c c c c c} 
 \hline
 Model & Accuracy & F1 & Avg. $R^2$ (Independence) & Rep2Label Accuracy \\[0.5ex] 
 \hline
Baseline Model & 89.8\% & 0.90 & (0.51, 0.1) & 94\%\\ 
RR Model & 94.5\% & 0.94 & 0.018 & 57\%\\
 P-Wave Model & 89.7\% & 0.90 & -0.082 & 58\%\\
 All Model & 92.1\% & 0.92 & -0.007 & 52\% \\
 \hline
\end{tabular}
\caption{Accuracy and F1 are reported on main task. Average $R^2$ and Rep2Label accuracy are for the auxiliary tasks. Baseline model is DNN without access to hand engineered features and without HSIC. RR model, P-Wave model and All Model are models with access to the respective hand-engineered features, along with DNN representations explicitly encouraged to be independent of the features.}
\label{table:ECG results}
\end{table}


\begin{figure} [ht]
\begin{tabular}{cc}
  \includegraphics[width=72mm]{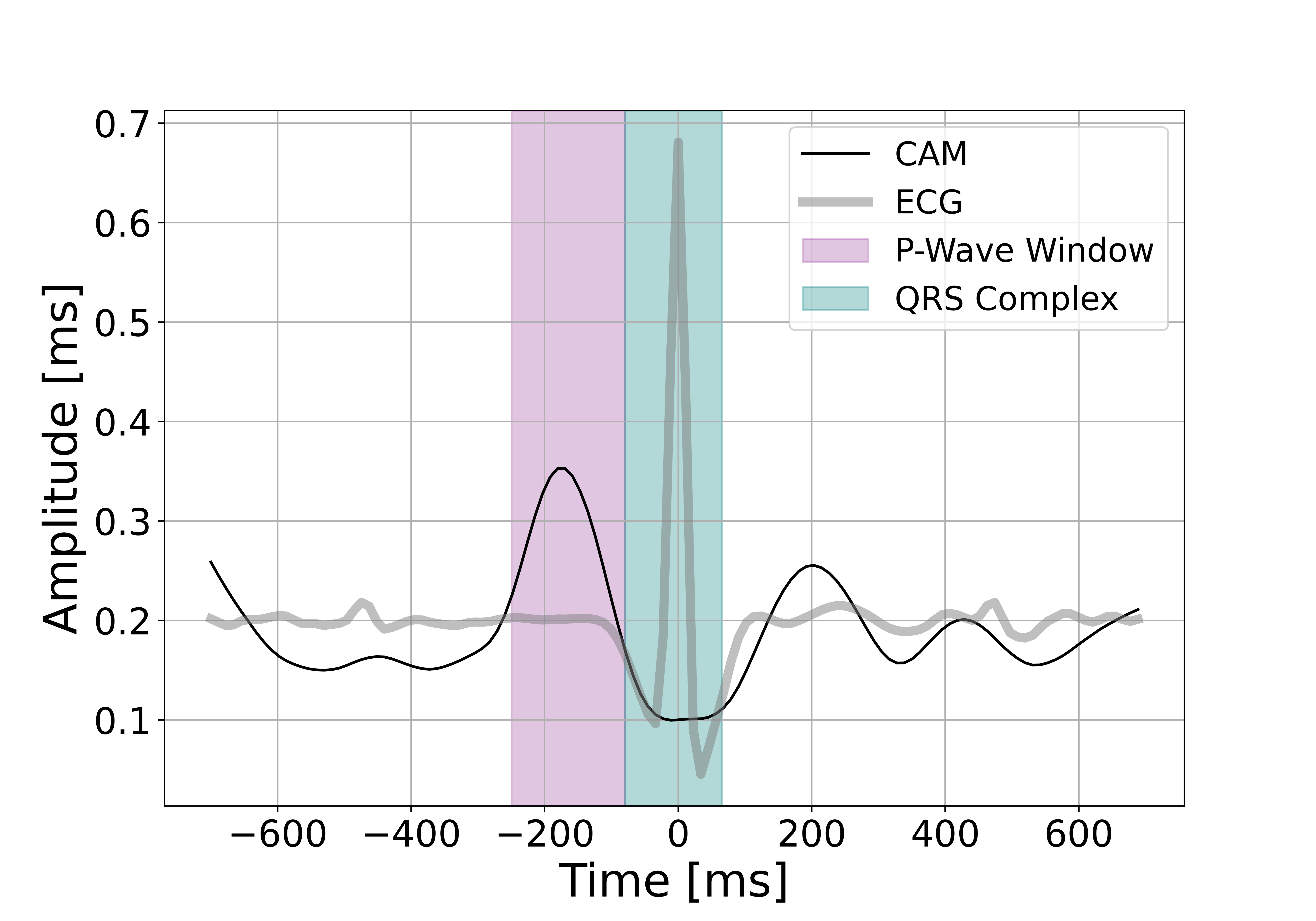} &   \includegraphics[width=72mm]{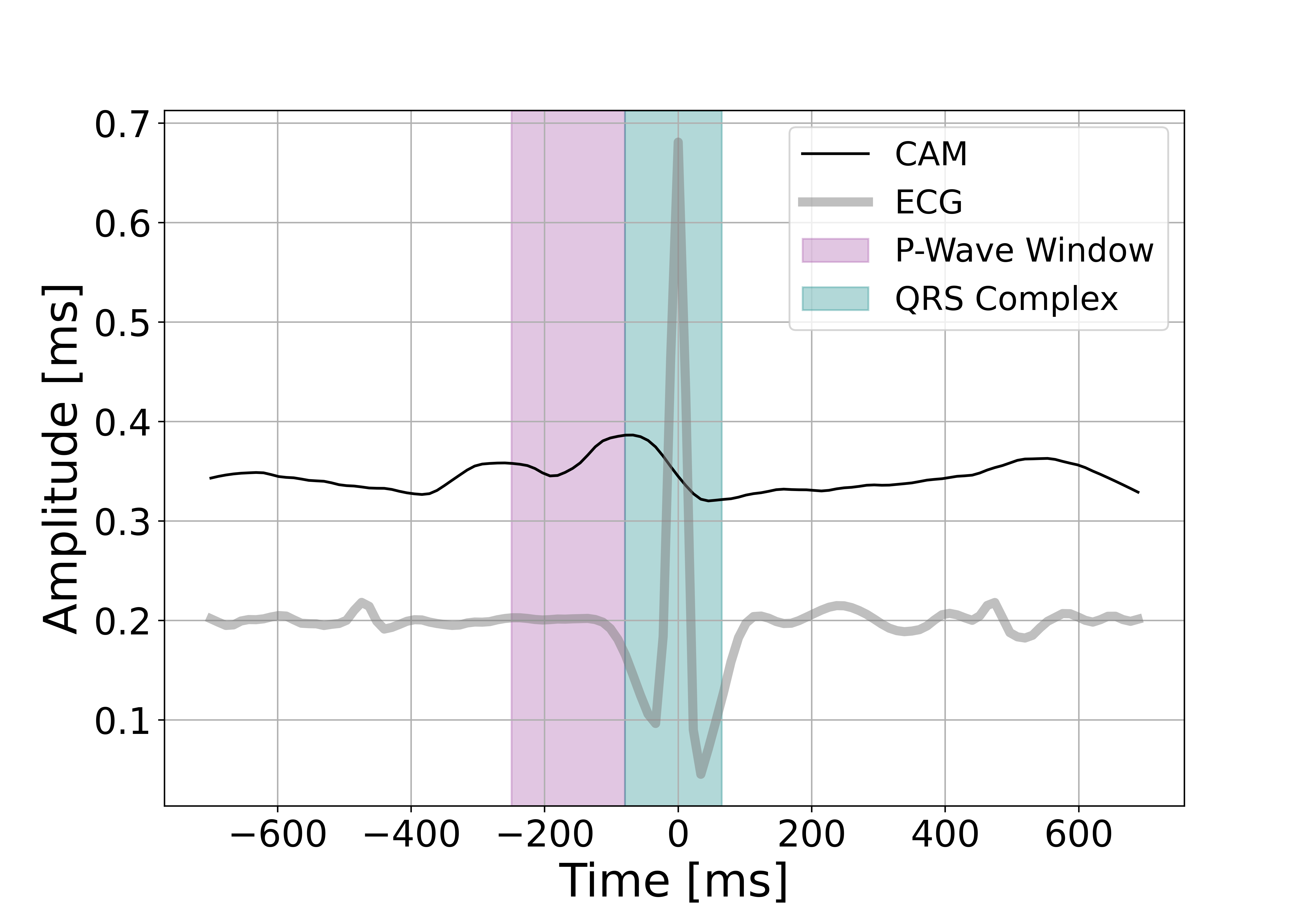} \\
(a) Baseline & (b) All \\[6pt]
 \includegraphics[width=72mm]{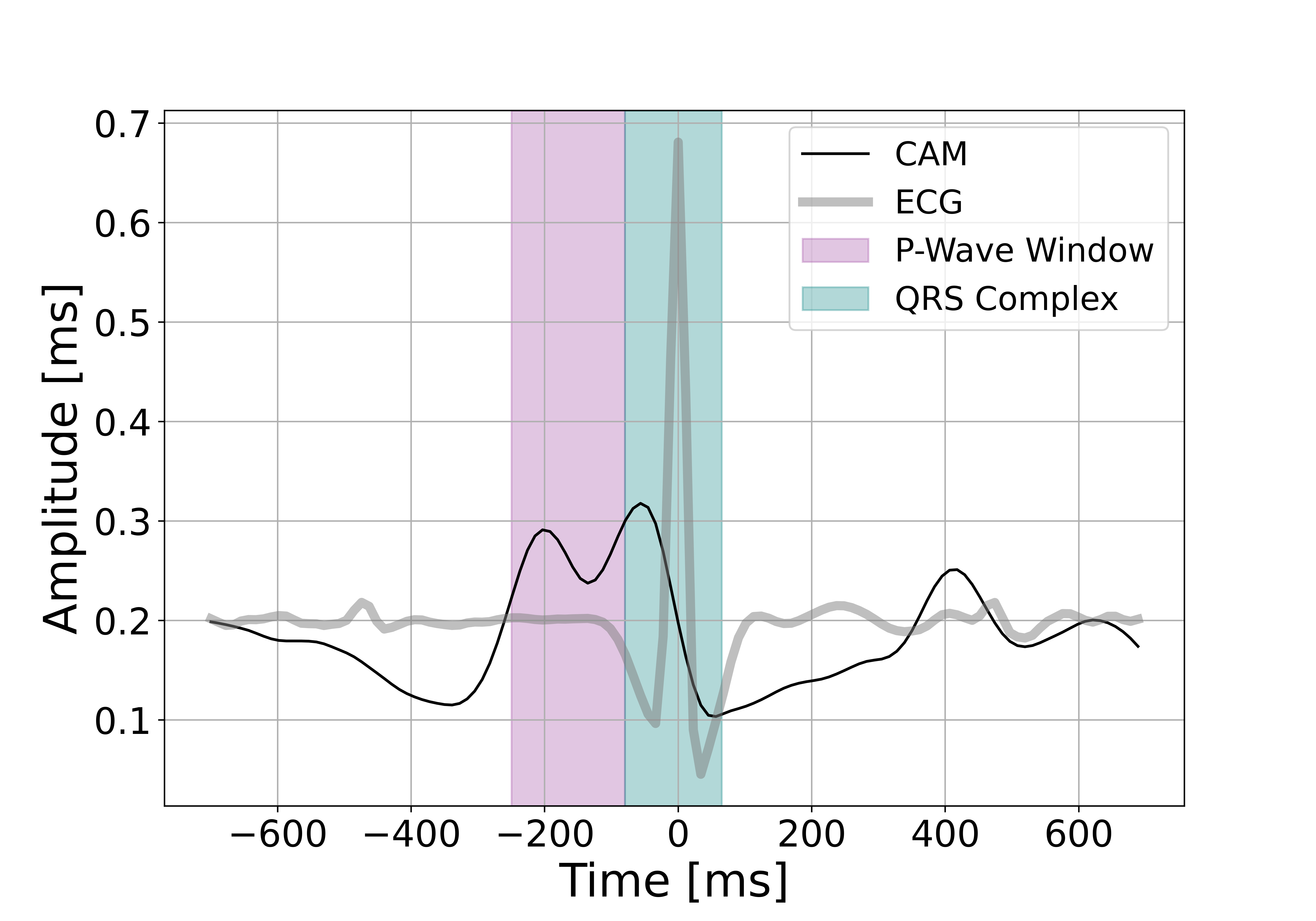} &   \includegraphics[width=72mm]{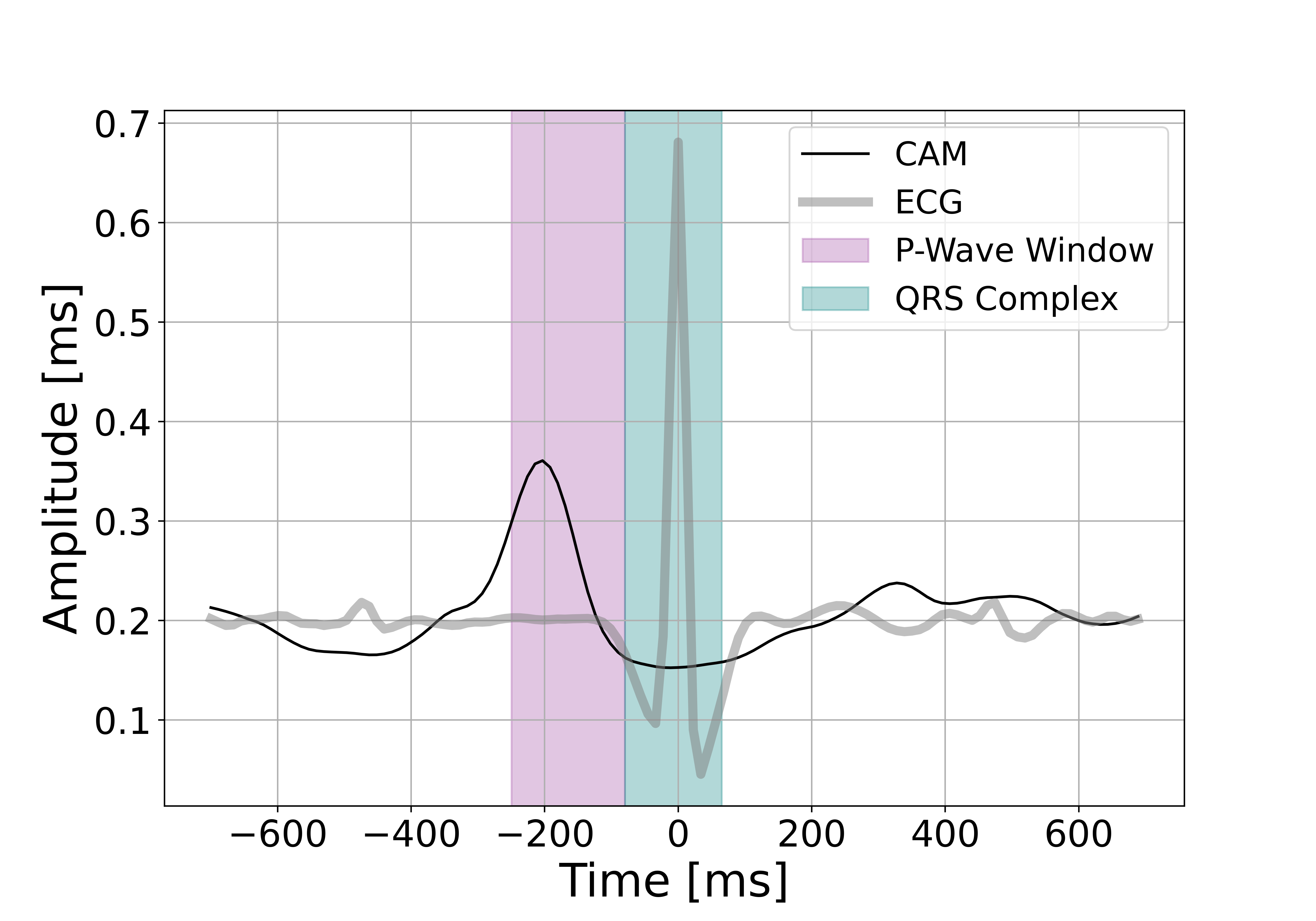} \\
(c) P-Wave constrained & (d) RR constrained \\[6pt]
\end{tabular}
\caption{Mean class activation map templates for all four models, superimposed over the mean QRS. The QRS and activation templates above are the result of averaging over 3839 detected QRS areas from all atrial fibrillation records in the test set.}
\label{fig:main_ecg_fig}
\end{figure}

Next, we examine the mean Class Activation Map template to observe the changes in the focus of the nets. In Figure \ref{fig:main_ecg_fig}, for the RR model (whose representation is encouraged \emph{not} to use R-peak related information), we observe an almost smooth activation with a clear peak which shifted toward the beginning of expected location of the theoretical P-wave window. In addition, in the P-Wave model's activation (i.e a model that is pushed to \emph{not} use P-wave-based features), it seems that the activation shifted toward the R-peak starting point. As one of the hallmark features of AF is the loss of regularity of the intervals between consecutive R-peaks, an increase in activation at the expected location of the R-peak seems to reflect the shift of the internal representation. According to paired t-tests performed on the activation within P-wave and QR windows as defined in medical literature: In the P-wave window, RR model CAM $>$ P-wave model CAM, $p<10^{-10}$. In the QR window, P-wave model CAM $>$ RR model CAM, $p<10^{-10}$. Further analysis of CAMs in the ECG signal is brought in Appendix \ref{appedix:noise}.

These results strengthen our belief that the network automatically learned features which temporally roughly correspond to the P-Wave and R-peaks. It also shows that our method can selectively force the network not to use pre-specified aspects of the signal, even if they are given as high-level features such as the RR and P-Wave features used here.

\subsection{Results: EEG Experiment}

Unlike the previous experiment, in this experiment we use only one set of features, which we call Frequency. Here however we experiment with two different types of landmarks for aligning the signals in the Class Activation Maps: The Slow Wave, which is extracted from the EEG signal itself, and rapid eye movements, which are extracted from the EOG signal. We emphasize that we do not use the EOG signal as input to any of our models, and its only use is in detecting the onset of REM used in creating the Class Activation Maps.

We first see in Table \ref{table:EEG Relevance} that the Frequency features are informative regarding the task at hand, with $69.5\%$ accuracy (for this specific auxiliary task the label set is balanced so chance performance is $50\%$).

We then see, in Table \ref{table:EEG results}, that the Independence task is successful: while the average $R^2$ predicting the frequency features from the representation of the baseline (unconstrained) network is 0.72, when using the Frequency model the latent representation becomes completely unable to predict the frequency features with average $R^2$ of -0.037.

Proceeding to the third auxiliary task, we examine how predictive of the label are the learned representations when standing on their own. Here we see that, in contrast to the ECG task, the representation learned to explicitly not use Frequency information does not lose anything in accuracy. 


On the main task, the Frequency model (i.e a model that is pushed to \emph{not} use frequency-based features, and then has the frequency features concatenated) achieves similar performance to the Baseline Model. 
Therefore, while the representation learned in the Frequency Model does not contain any information allowing the prediction of the frequency features, it still retains everything needed for a good prediction of the relevant steep stage.
The fact that the hand-crafted frequency features are less informative than they are in the previous experiment may have contributed to this outcome, but more importantly, there exists local features which are predictive to the label and yet independent of the frequency features which are global in nature. 

\begin{table}[h!]
\centering
\begin{tabular}{c c c c c c} 
 \hline
 Feature set & Accuracy & F1  \\[0.5ex] 
 \hline
 Frequency Features & 69.5\% & 0.64 \\
 \hline
\end{tabular}
\caption{EEG Relevance.}
\label{table:EEG Relevance}
\end{table}

\begin{table}[h!]
\centering
\begin{tabular}{c c c c c c} 
 \hline
 Model & Acc. & F1  & Avg. $R^2$ (Independence) & Rep2Label Acc. \\[0.5ex] 
 \hline
Baseline Model & 90.9\% & 0.90  & 0.72 & 90.5\%\\ 
Frequency Model & 90.2\% & 0.87  & -0.037 & 90.3\%\\
 \hline
\end{tabular}
\caption{Accuracy and F1 are reported on main task. Average $R^2$ and Rep2Label accuracy are for the auxiliary tasks. Baseline model is DNN without access to hand engineered features and without HSIC. Frequency Model is a model with access to the hand-engineered features, along with DNN representations explicitly encouraged to be independent of the features.}
\label{table:EEG results}
\end{table}

We now turn to examine the Class Activation Maps of the models with respect to the two choices of landmarks. 

First, we attempt to evaluate whether the activation related to the frequency features is indeed removed.
We do so by observing slow waves: While the Baseline Model's activation clearly focuses on slow waves locations, imposing independence in the Frequency model yields a very noisy activation around these locations. This is yet another indication for the reduction in frequency related information induced by our method, when the hand-engineered features are frequency related features. 

\begin{figure}
\begin{tabular}{cc}
  \includegraphics[width=75mm]{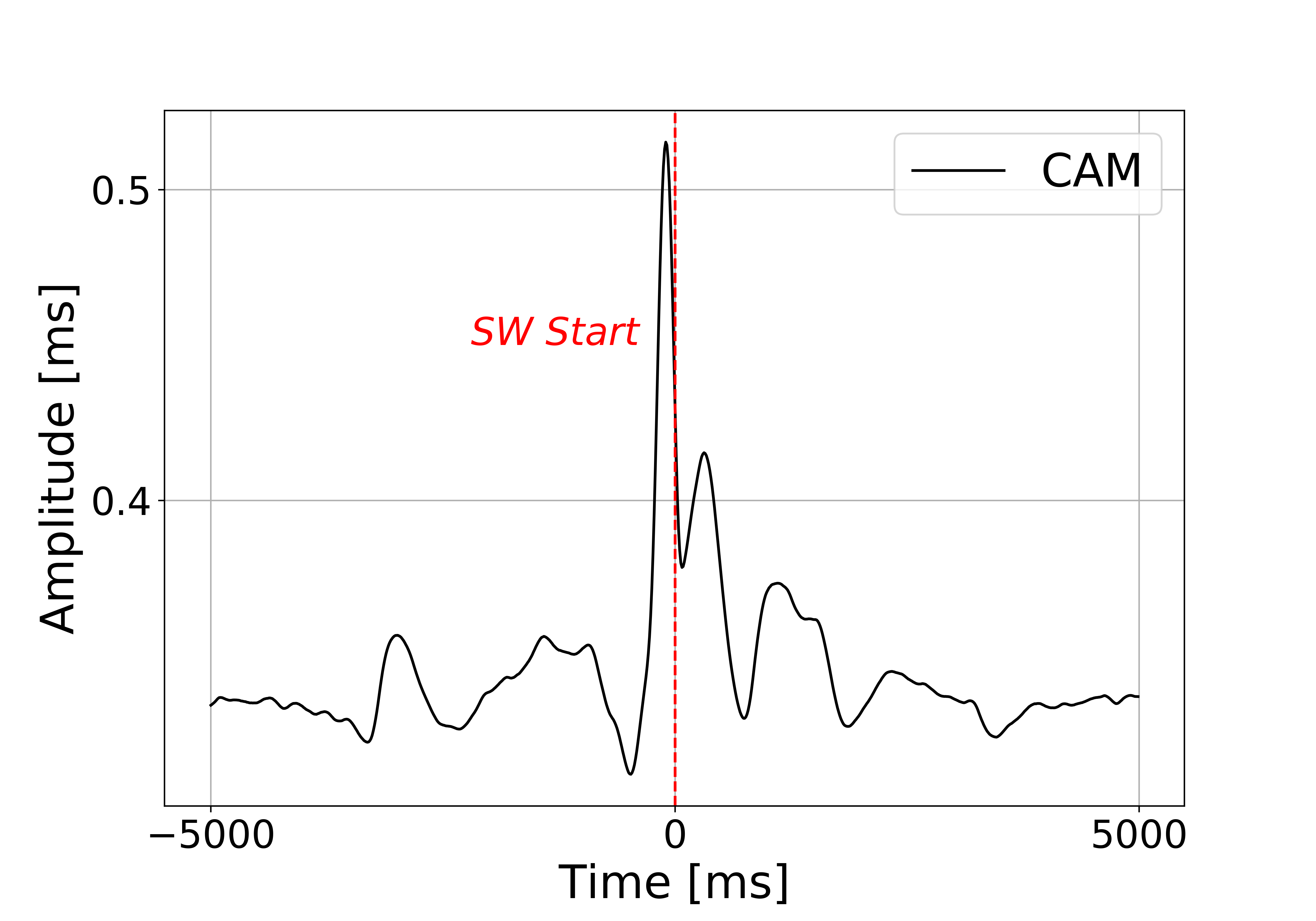} &   \includegraphics[width=75mm]{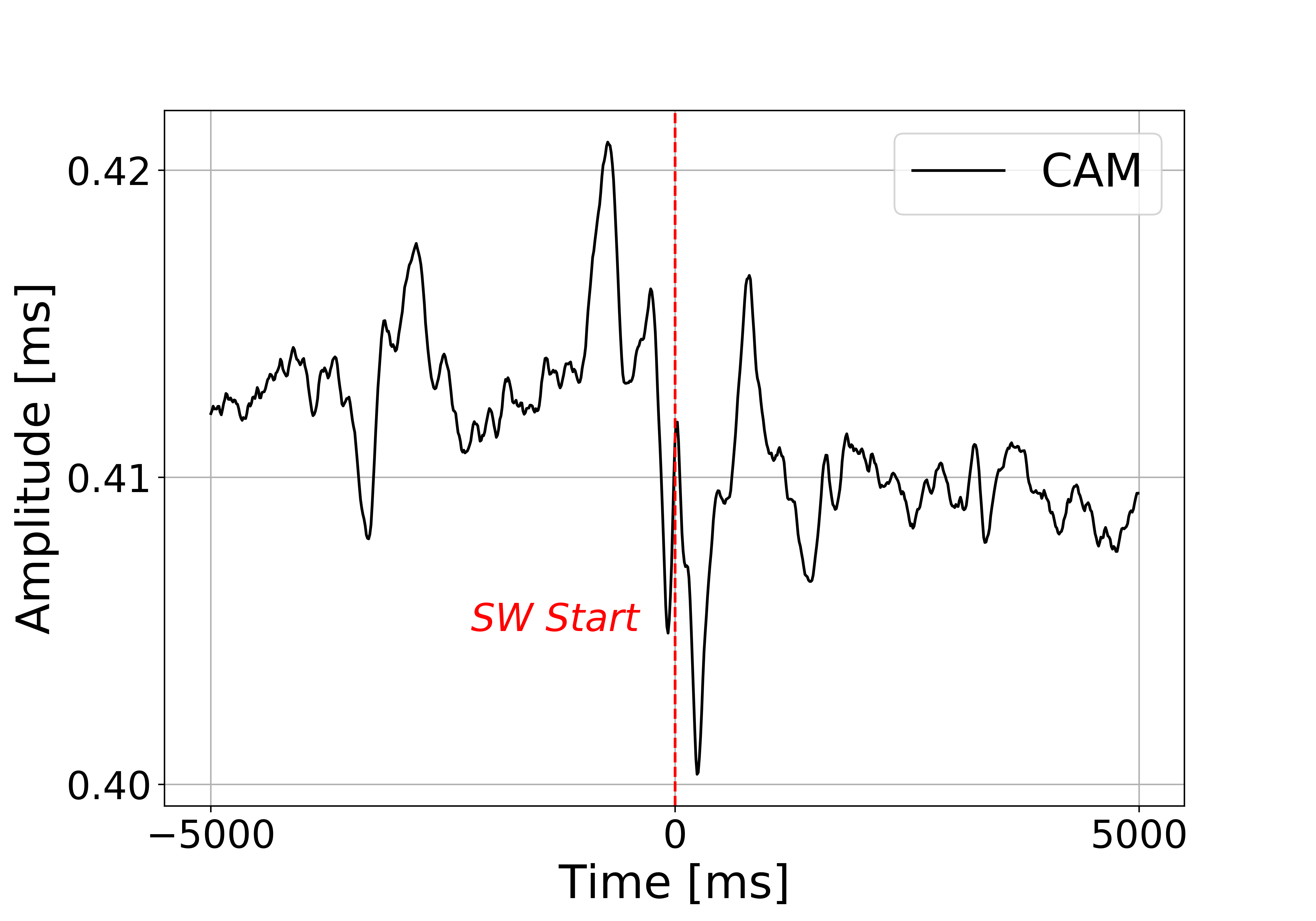} \\
(a) Baseline (SW alignment) & (b) Frequency Model (SW alignment) \\[6pt]
  \includegraphics[width=75mm]{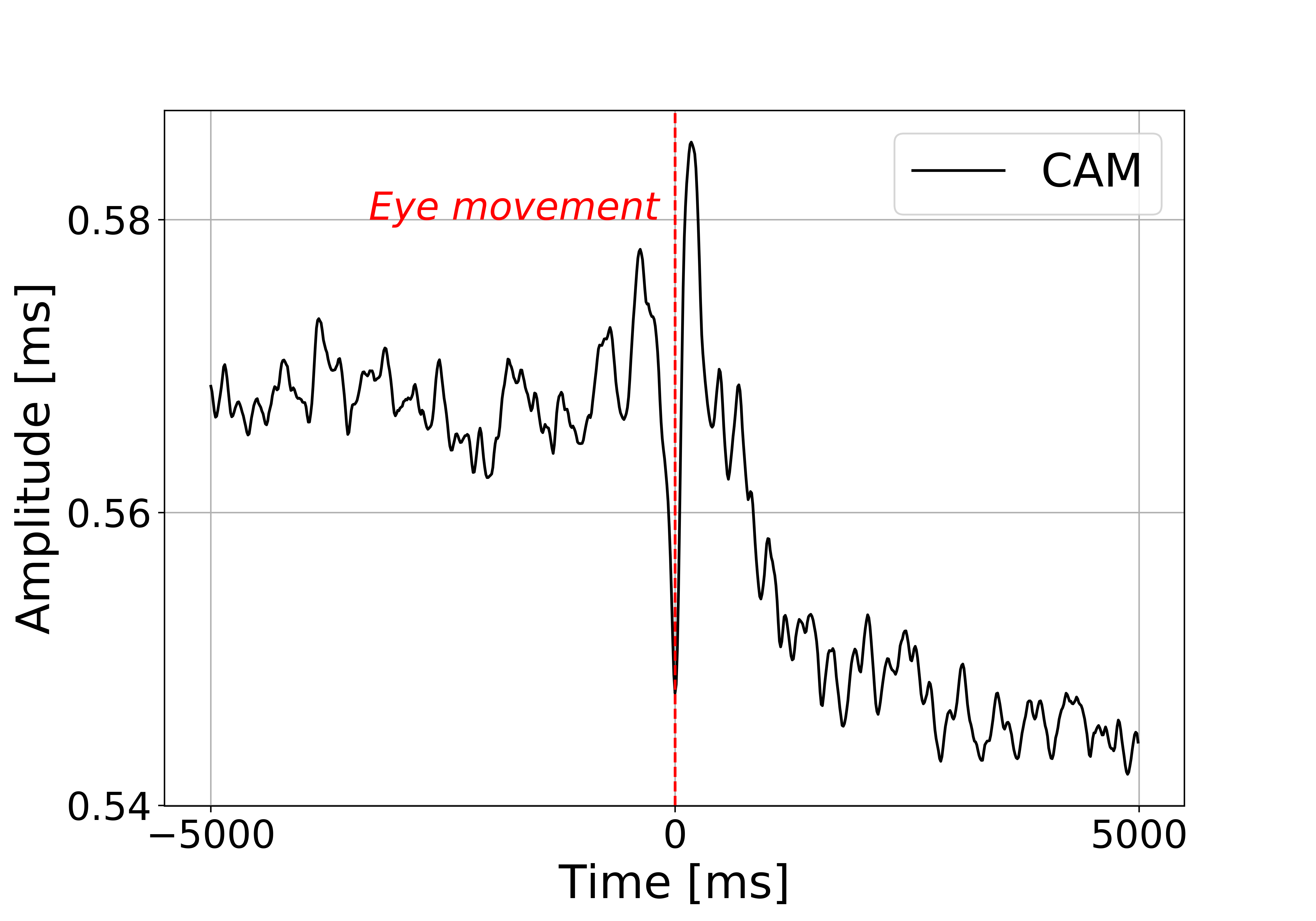} &   \includegraphics[width=75mm]{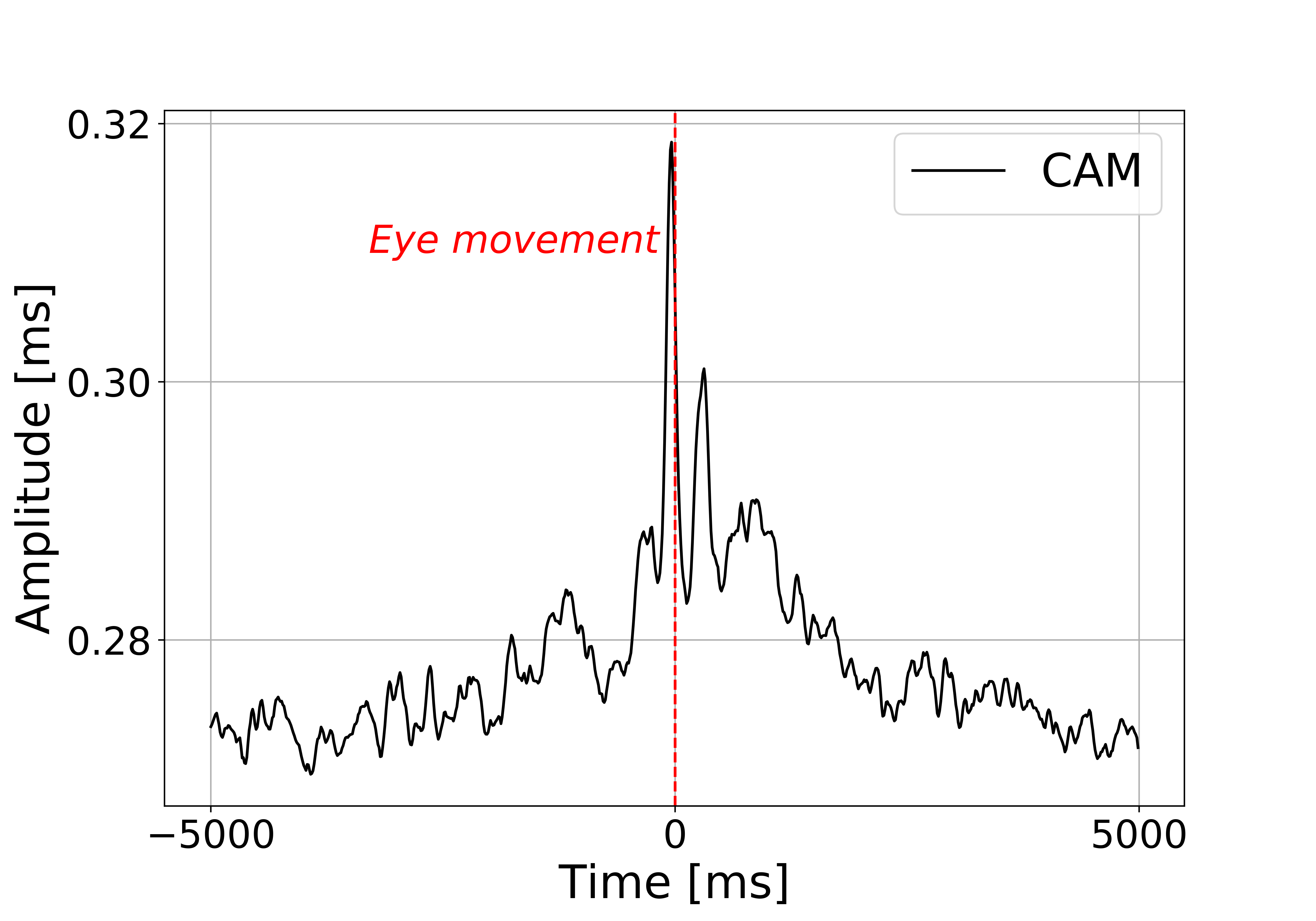} \\
(c) Baseline (eye movement alignment) & (d) Frequency Model (eye movement alignment) \\[6pt]
\end{tabular}
\caption{Mean activation template of slow waves locations (a), (b) and rapid-eye-movement locations (c), (d). Slow waves mean template is the result of mean over the CAM of 37327 slow wave locations detected over all the test set's NREM records. The rapid eye movement template is the mean CAM of 5556 rapid eye movements detected over all the test set's REM records.}
\end{figure}\label{fig:rem-activation}


Finally, we explore the potential of our proposed method as a means to test hypotheses. As noted above, the excellent performance of the Frequency model alludes to the existence of other features that might aid in classification of REM traces.  Specifically, we test the conjecture that the EEG traces might contain signatures of eye movements that could enable differentiating  between REM and non-REM sleep stages. Initially, we naively examined the activation over rapid eye movement locations (as extracted from an EOG trace available in the data-set). If we had a clear peak over these locations in examining the Baseline Model's activation, it could have served as additional evidence to their relevance for REM detection. However, as depicted in \ref{fig:rem-activation} (c), the activation is very noisy. On the other hand, after inducing independence and reducing redundancy from frequency features we get a clearer picture: The activation of the Frequency model yields a distinguishable peak that cannot be confused with noise, indicating the model used signatures of rapid-eye-movement for REM detection. We further analyze this pattern in Appendix \ref{appedix:noise}.



\section{Related Work}
The goal of augmenting neural representations with external feature spaces, with or without inducing independence, has been studied to some extent in various domains. Most of these works have attempted to either remove certain biases from the learned representation, or to improve its robustness in domain adaptation tasks.

In the computer vision domain, \citet{kim2019learning} propose a mutual information regularization term that encourages the network to 'unlearn' a predefined bias attribute. \citet{wang2019balanced} employ an adversarial approach to remove a protected attribute (e.g. gender) from the input space, yielding a ``debiased'' dataset. \citet{singh2020don} propose a loss term that penalizes overlap between class activation mappings of two co-occurring classes, mitigating spurious correlations. While they do succeed in learning a more robust representation of their target classes, their process still requires having prior knowledge about the label co-occurrence. \citet{atzmon2020causal} model object-attribute pairs by means of the intervention that has generated the image. They also apply HSIC to induce independence between object and attribute embeddings.

In natural language processing, there is a long tradition of injecting external knowledge to models. For example, \citet{kiperwasser2016simple} aims to extract a complementary representation for a known hand-crafted feature set. The boosted performance presented in the above work suggests that complementary information is indeed learned, but there might be redundancy between this complementary representation and the hand crafted features as no notion of independence is enforced. \citet{elazar2018adversarial} adversarially remove demographic attributes from a text corpus.

Building efficient interpretable systems is one of the major obstacles to deployment of machine learning models in healthcare \citep{tonekaboni2019clinicians}. Some studies have attempted to interpret the decisions made by their black-box models. For 2D imaging data, \citet{rajpurkar2017chexnet} inspect class activation mappings of their derived model to gain trust in its predictions. For 1D ECG data, \citet{goodfellow2018towards} inspect neural activations and find correlates between decision rules of experts and deep networks.

In general, while methods exist for learning a classifier while ignoring predefined aspects of the input, we wish to emphasize that our goal is slightly different: we wish to learn ``over and above'' the information extant in the predefined features. In a sense we wish to fit the residual left after predicting the label with the known feature. That is distinct from the methods such as that by \citet{kim2019learning} above which focus on ignoring the predefined features altogether. 

\section{Discussion}  \label{Discussion}


In this work we presented a new method for exploring to what degree Deep Neural Network representations replicate existing knowledge in terms of feature engineering for physiological signals. This method can further be used to discover new knowledge embedded within the representations of neural networks, by removing the traces of existing features from the learned representation enabling greater focus on potentially novel features.

After validating our method indeed succeeds in learning representations which are independent and non-redundant with respect to given hand engineered features, we make two further contributions. 
First, we give evidence that the success of DNN methods in classifying AF from ECG signals relies mostly on the DNN learning features that are very close to well-known hand-engineered features. This is both evidence to the strength of DNNs as well as their limitations. On the one hand, it is impressive to find that DNNs can automatically discover what took years of human research. On the other hand it is perhaps disappointing to see that current DNN models cannot tell us much new about the ECG signal inasmuch as it relates to AF. We further show that our method could in principle be used to discover new signals, by showing how, once the RR features are removed, the activation focuses strongly on the P-Wave section, discovering its importance for classifying AF.
Second, we show in EEG signals that our method can help focus the attention of researchers on potentially interesting signal attributes. Simulating a situation where a researcher has a hypothesis about the existence of a signature of a physiological event (eye movement) in the EEG trace \emph{and} the ability to locate such events in a separate signal (EOG in our case), we show how applying our method can focus the activation signal around a hypothesized event, helping the researchers validate their hypothesis.

\paragraph{Limitations}
Our study suffers several limitations: the framework we propose is relevant mostly to local features due to the nature of the Class Activation Maps. Class activation mappings require a specific kind of architecture: feature maps must directly precede softmax layers, so it is only applicable to a particular kind of CNN architecture performing global average pooling over convolutional maps immediately prior to prediction.  
Furthermore, the real challenge of feature discovery is not nearly solved by our method. A domain expert needs to translate the activation templates and come up with interpretable functions or scientific understanding that capture the exposed pattern.

\paragraph{Future Work}
We wish to extend our work to apply to global features more easily. More broadly, we would like to be able to interface with a much broader spectrum of methods for explaining and understanding DNNs, instead of relying heavily on CAMs. We also believe this approach can be extended to other physiological signals and possibly to other modalities such as images to drive scientific discovery and hypothesis testing. 


\acks{
We thank Daniel Greenfeld for comments that improved this work. This research was partially supported by the Israel Science Foundation (grant No. 1950/19).
}

\bibliography{main}

\appendix
\section{Noise Analysis}
\label{appedix:noise}
\begin{figure} [htbp]
\begin{tabular}{cc}
  \includegraphics[width=65mm]{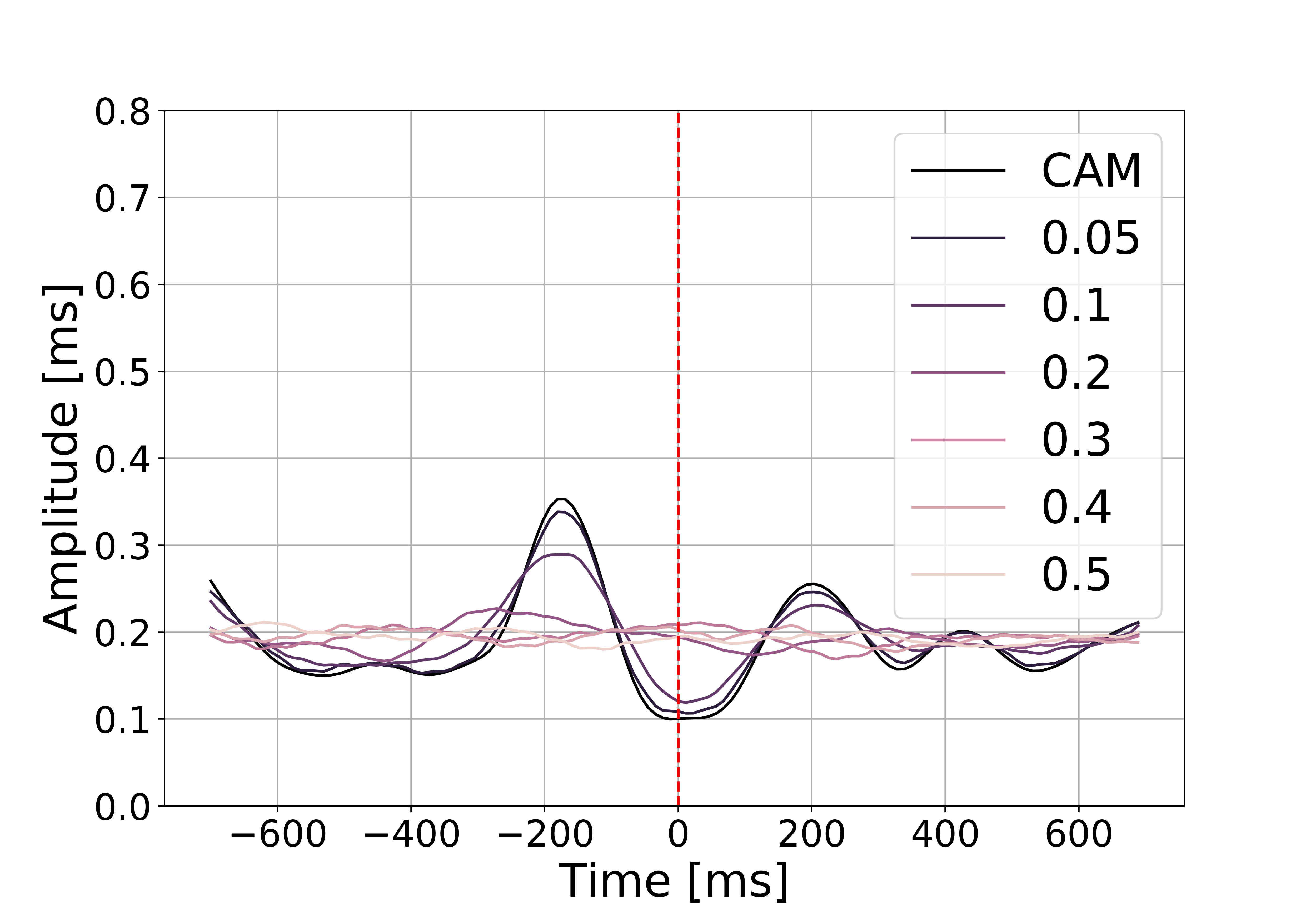} &   \includegraphics[width=65mm]{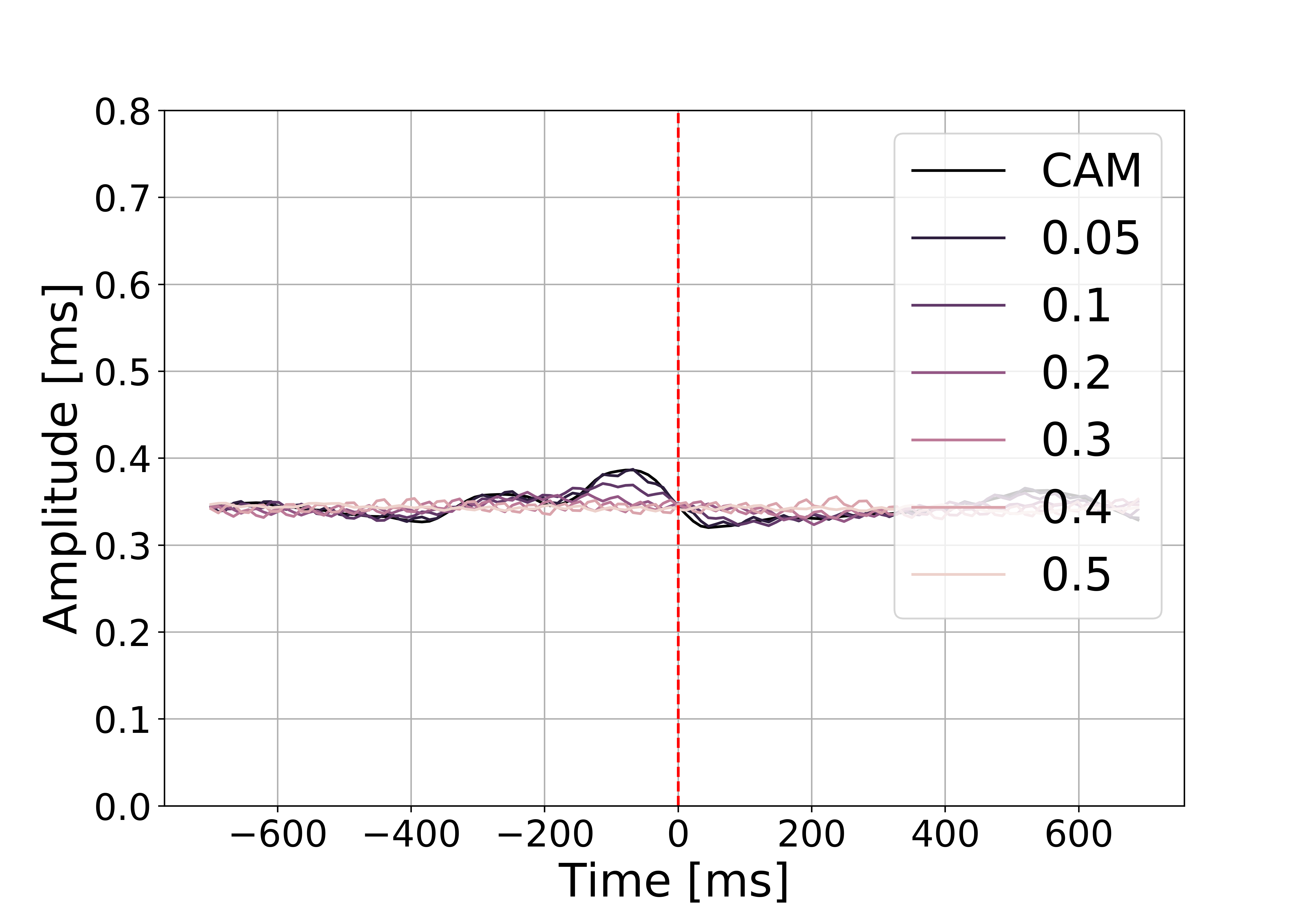} \\
(a) Baseline & (b) All \\[6pt]
 \includegraphics[width=65mm]{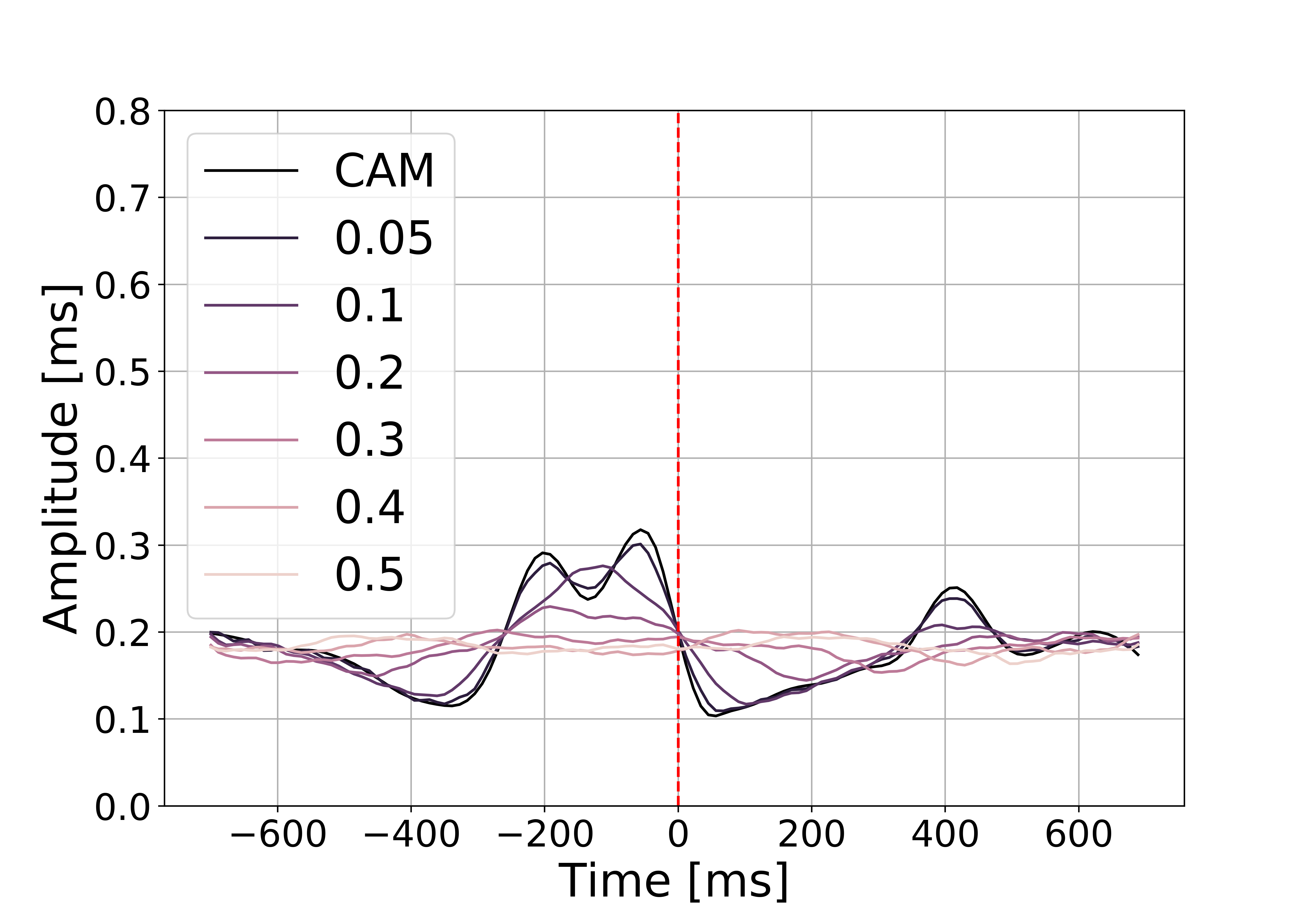} &   \includegraphics[width=65mm]{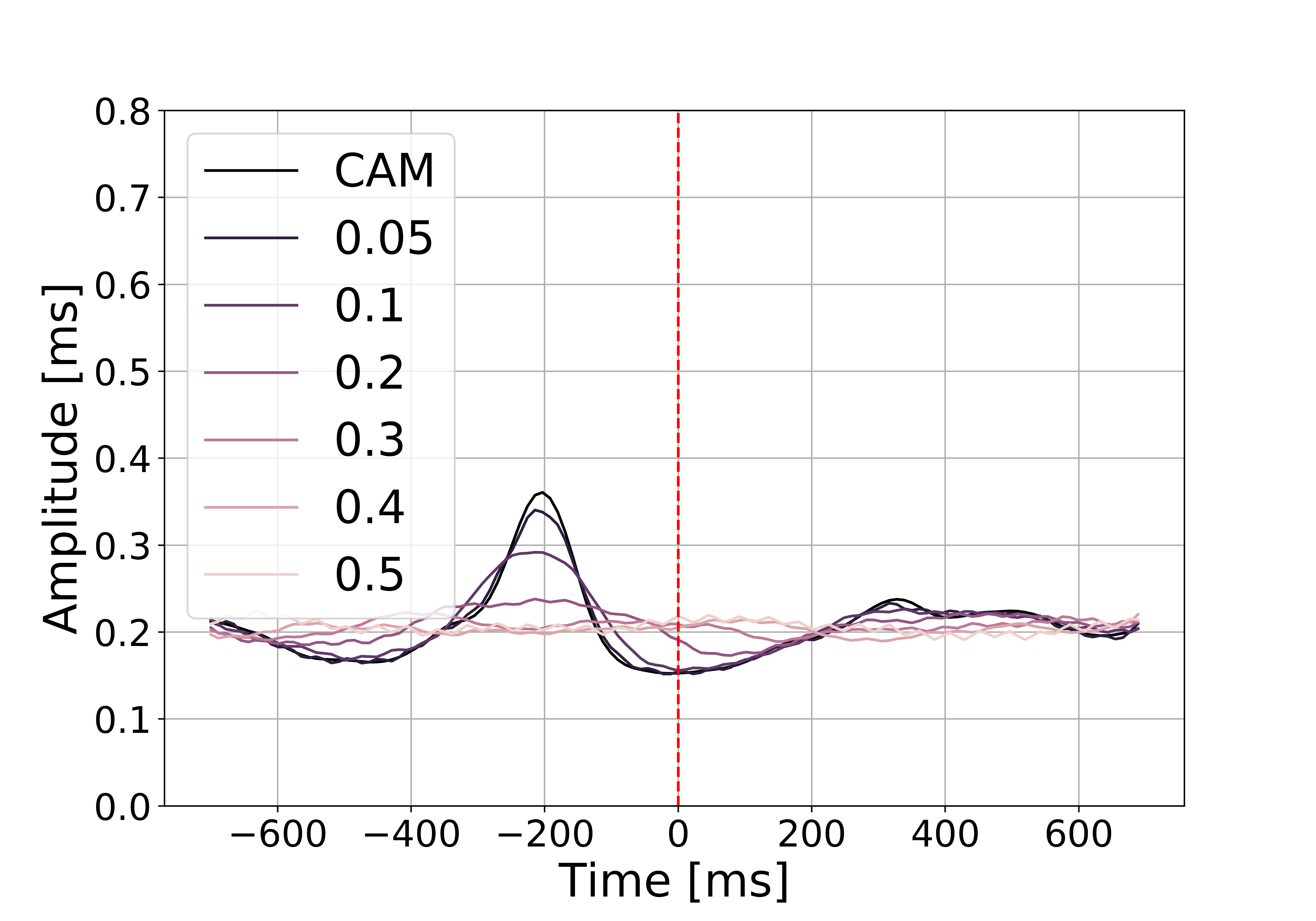} \\
(c) P-Wave constrained & (d) RR constrained \\[6pt]
\end{tabular}
\caption{Noise analysis of the ECG experiment}
\label{fig:stat_ecg}
\end{figure}

In order to obtain additional validation for the activation template analysis we tested the behavior of the mean template activation with increasing noise.
Instead of averaging templates centered around a detected event such as R-Peaks, each event location was corrupted with increasing noise intensity. For each corruption intensity, noise was sampled uniformally in $[0, intensity]$ and was added to the detected event. This way, the templates are not perfectly aligned to the true events. 
Figure \ref{fig:stat_ecg} shows this analysis on the results of the ECG experiment. Compared to the uncorrupted analysis of Figure \ref{fig:main_ecg_fig}, here the meaningful activation patterns are nullified with increasing noise intensity, indicating that the network has indeed learned to focus its attention on the QRS area.

\begin{figure}[htbp]
\begin{tabular}{cc}
  \includegraphics[width=65mm]{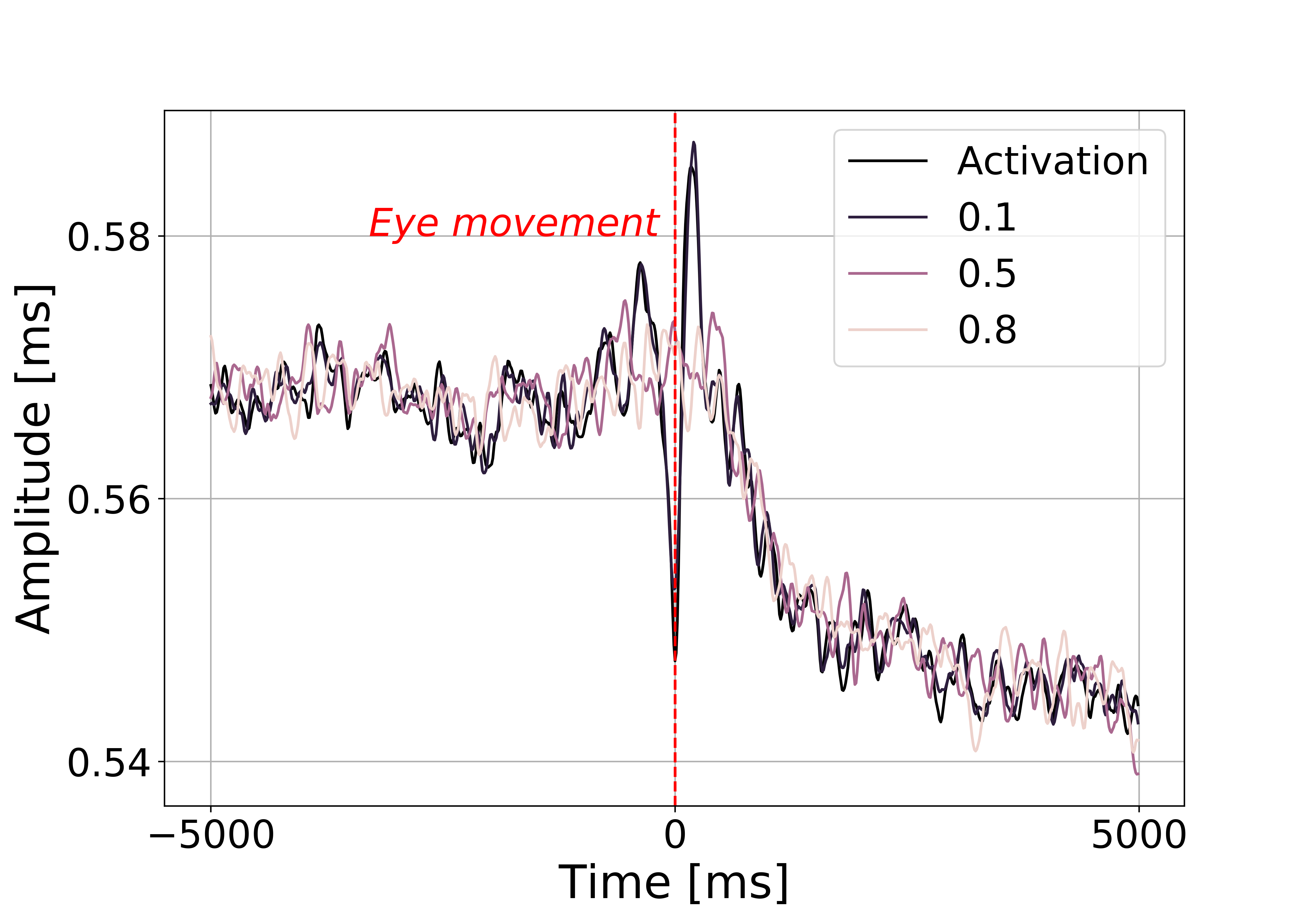} &   \includegraphics[width=65mm]{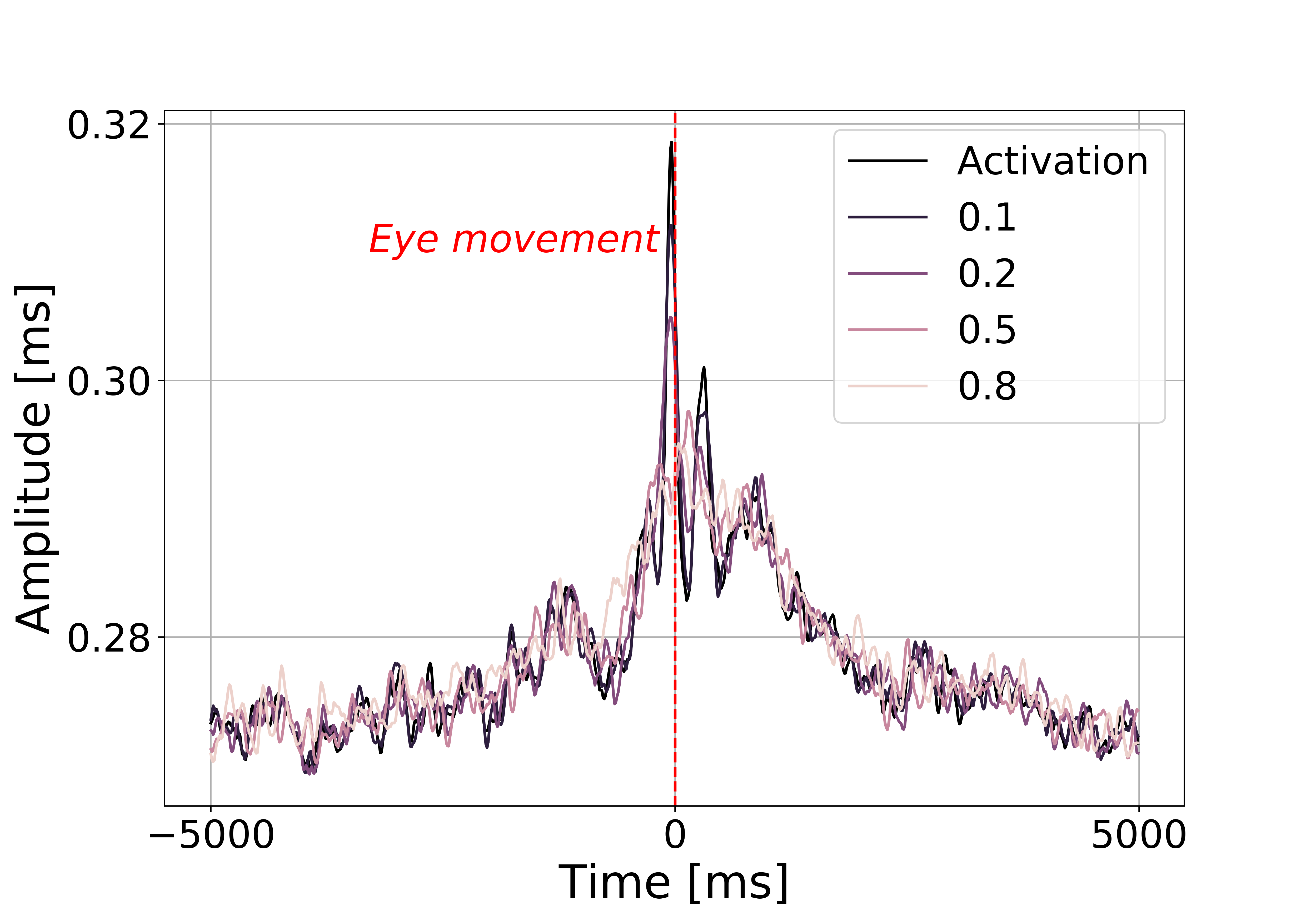} \\
(a) Baseline & (b) Frequency Model \\[6pt]
\end{tabular}
\caption{Noise analysis for the EEG experiment}
\label{eeg_noise}
\end{figure}

We apply the same analysis to the EEG experiment, as shown in Figure \ref{eeg_noise}.


\section{Additional EEG Analysis}
\label{appedix:additional-eeg-analysis}

\begin{figure}[htbp]
\begin{tabular}{cc}
  \includegraphics[width=65mm]{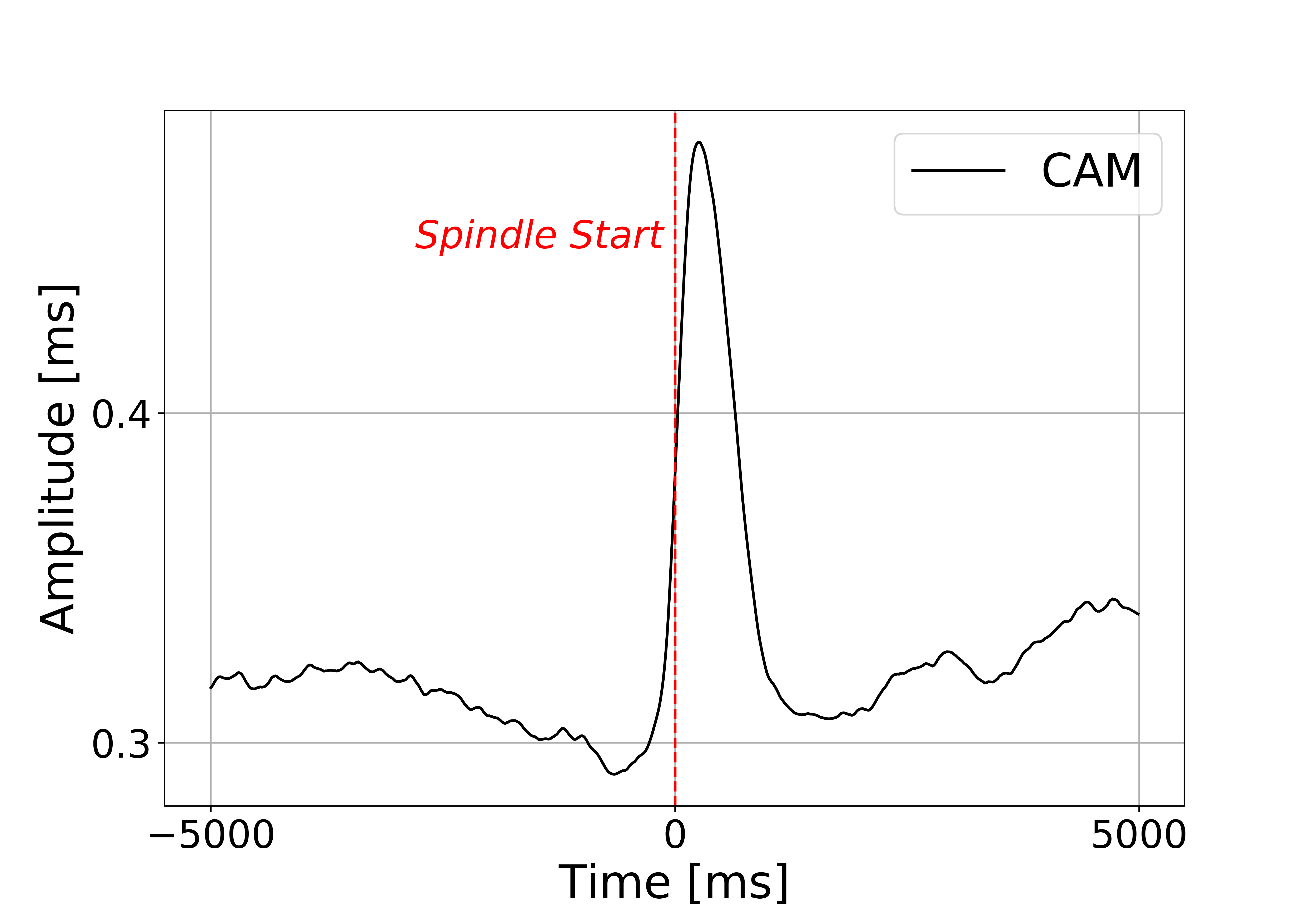} &   \includegraphics[width=65mm]{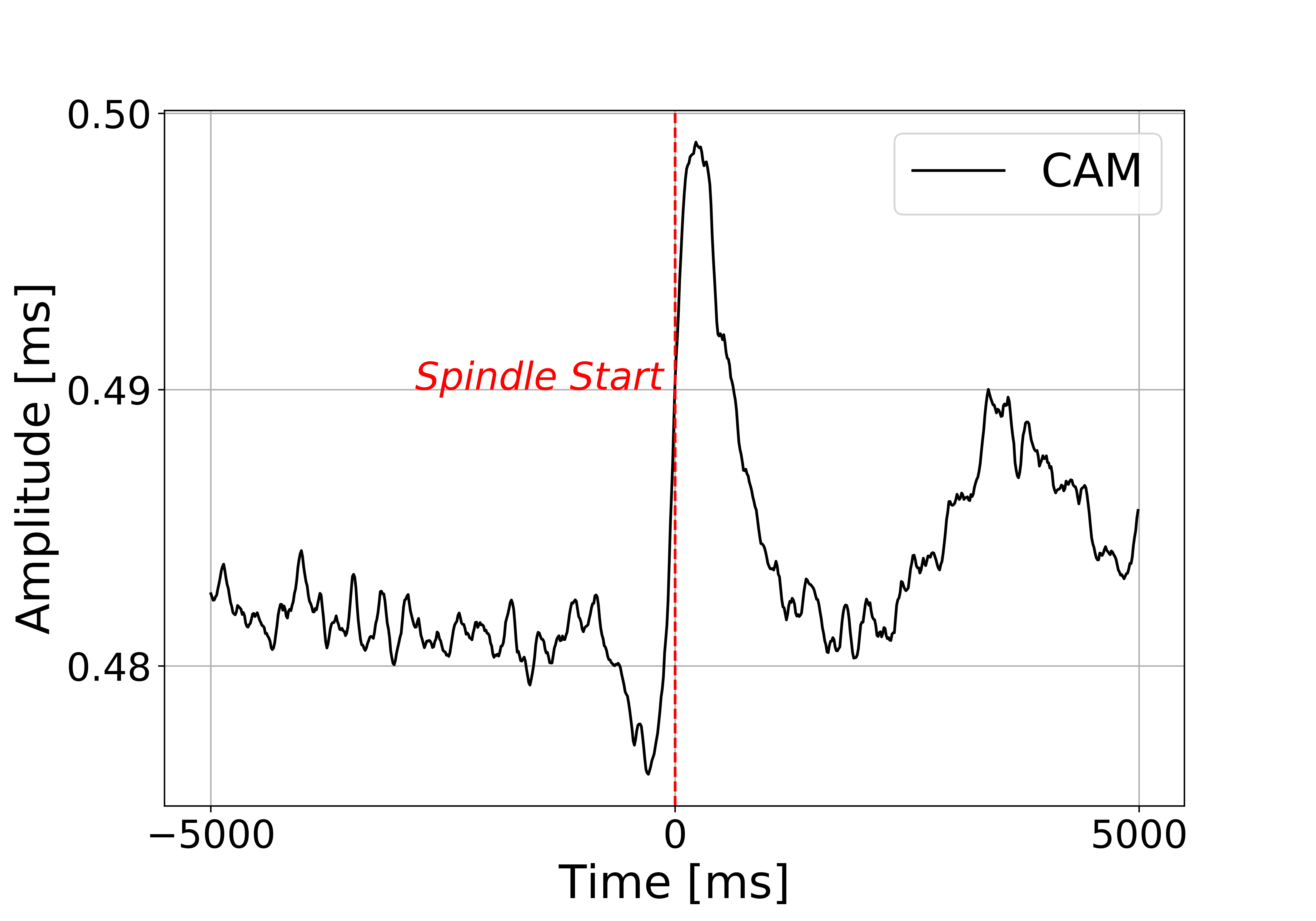} \\
(a) Baseline & (b) Frequency Model \\[6pt]
\end{tabular}
\caption{Mean activation template of sleep spindles locations. Mean over 19219 spindles locations detected over all test set's NREM records.}
\label{fig:spindles}
\end{figure}

As another hypothesis test, we examine the class activations around events of sleep spindles. Figure \ref{fig:spindles} shows that baseline activation over sleep spindles is significant and remains so also in the frequency-constrained model. As opposed to slow waves, sleep spindles are not defined exclusively by the their frequency content - they are short and local bursts of activity (lasting 0.5-1.5 seconds) \citep{paruthi2016recommended}. Hence they exhibit characteristics of local morphology that is not captured by the global frequency features. This analysis shows that the proposed framework does not simply obstruct the activation of any local features.

\section{Specifications about the models} 
\label{appedix:model_details}
In this section we provide the technical details required to reproduce the results 
\subsection{Main Task}
\label{appendix:main_details}

We base our main model on the architecture proposed in \citet{goodfellow2018towards} with some modifications required for the introduction of hand-engineered features. This model is a 12-layer convolutional neural network inspired by WaveNet \citep{oord2016wavenet}. In all convolutions the kernel size is fixed to be 3, and unless noted otherwise, the number of output channels is 128.

The first block is a convolution with a stride of 2, reducing the input dimension by half. Following are 9 identical residual blocks, as described in \citet{oord2016wavenet}, each with a doubled dilation factor (starting at 2 and ending at 512). Skip connections from all residual blocks are summed and passed to the next block, which is comprised of the two last convolutions, with 256 output channels for the first and 512 for the second. Each convolution precedes a ReLU activation, max-pooling and dropout (with probability 0.3). At this stage, the temporal dimensionality of the latent representation is still half of the input size, which is key for meaningful extraction of class activation mappings. Next, the temporal dimension is averaged by a global average pooling (GAP) layer, resulting in an output of size 512 which is then batch-normalized. This output is regarded to be the network's latent representation and is denoted with $g$. 
After it is concatenated along the external feature set $f_i$, this unified representation is passed to a fully connected linear layer yielding the class logits.

The hyper-parameters for each of the two tasks are selected after training baseline models (with no hand-crafted features) to achieve performance comparable to state of the art on the respective classification tasks, and are detailed in the following.
 
\begin{figure}[htbp]
  \centering 
  \includegraphics[width=0.5\textwidth]{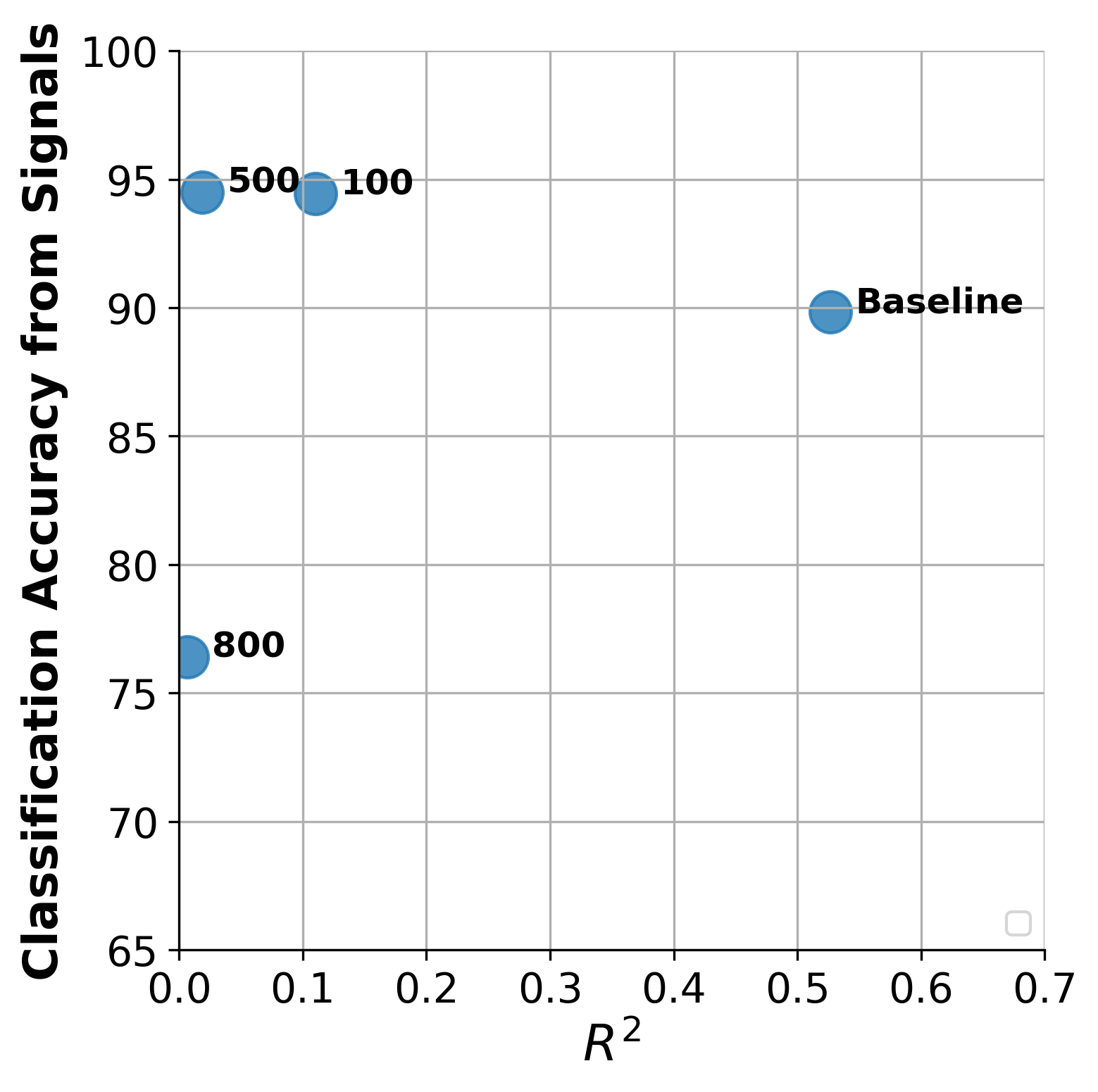} 
  \caption{RR model selection: The tuning of the $\lambda$ parameter for each constrained model is done according to success on the main task vs. $R^2$ in predicting the hand crafted features. In this case $\lambda$ 500 is chosen}
  \label{fig:rr_model_selection} 
\end{figure}

For the ECG experiment, the model is trained with cosine annealed learning rate, starting from $10^{-3}$ and ending at $10^{-5}$. It is trained for 200 epochs with a batch size of 32. The HSIC scaling parameter, $\lambda$, is calibrated by measuring the models' performance in both objectives (classification and independence, see section \ref{Evaluation Approach}). Figure \ref{fig:rr_model_selection} demonstrates the model selection process. According to this figure, this parameter was set to 500, achieving both excellent classification performance and sufficient independence.

For the EEG experiment, the model is trained with cosine annealed learning rate, starting from $10^{-5}$ and ending at $10^{-7}$. It is trained for 60 epochs with a batch size of 32.

Due to significant fluctuations in the magnitude ratio of the objective function's components, a desirable model was not found using a fixed value for $\lambda$. To circumvent this, the model was allowed a warm start of 20 epochs with $\lambda=0$ (training only the classification objective). Then, this weighting parameter was set at each batch to keep a constant proportion of 0.25 between the classification loss and the HSIC loss. This choice met the requirements of sufficient classification performance and independence to the external feature set.

\subsection{Auxiliary Tasks}
\label{appedix:aux_details}
For all three tasks described in section \ref{Evaluation Approach} we used 2-layered fully connected networks with ReLU activations. These networks have two hidden layers with 128 units, and were trained for 40 epochs with 32 samples per batch and a learning rate of 0.0003. The two first tasks (relevance of hand-engineered features and independence between hand-engineered and DNN features) used the same train/validation/test split as the main task. The third task (label information in latent representation), however, was trained, validated and tested only on the main task's test set. Due to its relatively small sample size, model training was done via stratified 5-fold cross validation of this test set.

\end{document}